\documentclass[10pt,twocolumn,letterpaper]{article}

\usepackage[margin=0.75in]{geometry}
\usepackage[T1]{fontenc}
\usepackage{newtxtext}
\usepackage{newtxmath}
\usepackage{microtype}
\usepackage[hyphens]{url}
\usepackage{graphicx}
\usepackage[round,authoryear]{natbib}
\usepackage{caption}
\usepackage{amsmath}
\usepackage{amsfonts}
\usepackage{algorithm}
\usepackage{algpseudocode}
\usepackage{booktabs}
\usepackage{multirow}
\usepackage{array}
\usepackage{float}
\usepackage[section]{placeins}
\usepackage{titlesec}
\usepackage{xcolor}
\definecolor{linkblue}{RGB}{0,70,140}
\usepackage[
    colorlinks=true,
    linkcolor=linkblue,
    citecolor=linkblue,
    urlcolor=linkblue
]{hyperref}

\hypersetup{
    pdftitle={RIDGE: Re-Noising with Internal Dynamic Guidance for Image Editing},
    pdfauthor={Ruiliang Gong, Zhen Wang, Yanghao Wang, Long Chen}
}

\titleformat{\section}{\large\bfseries}{\thesection}{0.55em}{}
\titleformat{\subsection}{\normalsize\bfseries}{\thesubsection}{0.55em}{}
\titlespacing*{\section}{0pt}{2.0ex plus 0.4ex minus 0.2ex}{0.8ex}
\titlespacing*{\subsection}{0pt}{1.5ex plus 0.3ex minus 0.2ex}{0.5ex}

\newcommand{\ie}{\textit{i.e.}}

\newcommand{\best}[1]{\textbf{#1}}
\newcommand{\second}[1]{%
  \begingroup\fboxsep=1pt\fbox{\strut #1}\endgroup}
\newcommand{\third}[1]{\underline{#1}}
\newcommand{\oursrow}{}

\title{\Large\bfseries RIDGE: Re-Noising with Internal Dynamic Guidance for Image Editing}
\author{
\large Ruiliang Gong \quad Zhen Wang \quad Yanghao Wang \quad Long Chen\textsuperscript{*}\\[0.55em]
\small Department of Computer Science and Engineering\\
\small The Hong Kong University of Science and Technology\\[0.4em]
\footnotesize
\texttt{rgongac@connect.ust.hk}~$\cdot$~%
\texttt{zhenwang@ust.hk}~$\cdot$~%
\texttt{ywangtg@connect.ust.hk}~$\cdot$~%
\texttt{longchen@ust.hk}\\[0.35em]
\footnotesize\textsuperscript{*}Corresponding author
}
\date{}

\begin{document}

\makeatletter
\twocolumn[
\begin{@twocolumnfalse}
\maketitle
\begin{abstract}
Inversion-free flow-based image editing avoids latent inversion, but still requires a target-side state at every editing step. The widely used equal-displacement construction keeps the displacement between the noisy source state and the target-side state unchanged across noise levels. This is inconsistent with noising, under which the displacement between two clean states noised with the same noise level and noise sample should contract as the noise level increases. Thus, it can lead to overly aggressive updates at high noise levels. We introduce RIDGE: Re-Noising with Internal Dynamic Guidance for Image Editing, an inversion-free and training-free method that maintains the edited state as an evolving approximation to the unavailable clean target state. RIDGE re-noises this approximation using the same noise level and noise sample as the clean source state, allowing their noisy displacement to decrease naturally with increasing noise. Since the edited state initially contains limited target semantics, RIDGE further applies internal dynamic guidance during the early high-noise steps. A clean target state prediction guides the provisional edited state through a soft dynamic mask derived internally from the model, focusing guidance on regions that require modification without external segmentation or detection models. Experiments on two benchmarks using two backbones, SD3 Medium and FLUX.1-dev, show that RIDGE offers a favorable aggregate trade-off among source preservation, target alignment, and perceptual quality.
\end{abstract}
\vspace{0.8em}
\end{@twocolumnfalse}
]
\makeatother

\section{Introduction}

Recent flow-based text-to-image models~\citep{labs2025flux1kontextflowmatching,esser2024scalingrectifiedflowtransformers} have become strong generative backbones for text-based image editing. Given a source image with a source prompt, the task is to modify the image according to a target prompt that specifies the desired changes. The main challenge lies in balancing two objectives: faithfully realizing the intended edit while avoiding unnecessary changes to the source structure, identity, or background.

\begin{figure*}[t]
    \centering
    \includegraphics[width=\textwidth]{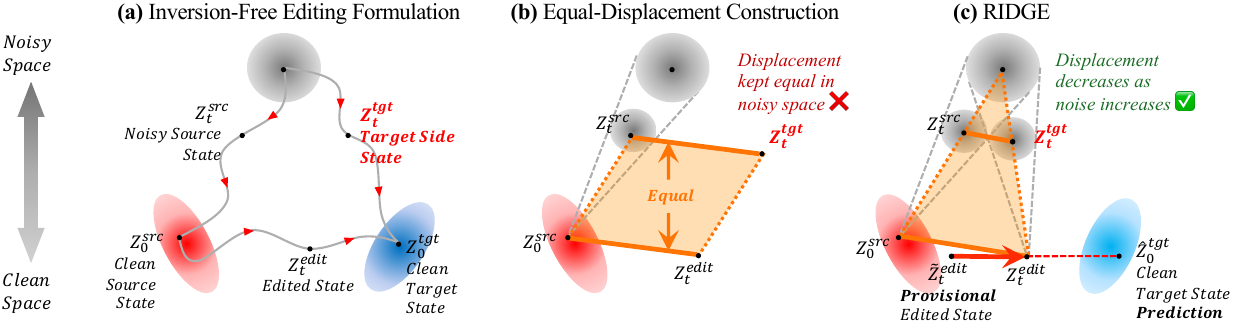}
    \caption{Target-side state construction in inversion-free editing. (a) A typical inversion-free formulation constructs a noisy source state from the clean source state and requires a corresponding target-side state for editing. (b) The equal-displacement construction keeps the noisy displacement equal to the clean displacement at every noise level. (c) RIDGE re-noises the edited state with the same noise level and noise sample as the clean source state, while internal dynamic guidance refines this approximation during early high-noise steps.}
    \label{fig:intro}
\end{figure*}

Early editing solutions are inversion-based methods: the source image is first inverted to a noisy latent state and then a new image is reconstructed under the target prompt~\citep{rout2024semanticimageinversionediting,wang2025tamingrectifiedflowinversion,xu2025unveilinversioninvarianceflow}. However, the inversion process can be slow and prone to reconstruction errors, motivating inversion-free methods to bypass it~\citep{kulikov2025floweditinversionfreetextbasedediting,kim2025flowaligntrajectoryregularizedinversionfreeflowbased,beaudouin2026deltarectifiedflowsampling,jiang2026snreditstructureawarenoiserectification}. As shown in Fig.~\ref{fig:intro}(a), a typical inversion-free formulation represents the source image in latent space as the \emph{clean source state}. The editing process starts from this clean source state and aims to reach the final edited result. At each editing step, the \emph{edited state} denotes the current intermediate clean state along this process, and is iteratively updated toward the final edited result. To compute this update, the source-prompt velocity is predicted from a \emph{noisy source state}, obtained by noising the clean source state at the current noise level. Likewise, the target-prompt velocity is predicted from a corresponding \emph{target-side state}. The difference between these two velocities determines the update direction of the edited state.

Therefore, constructing a noisy source state and a target-side state is required at every editing step. The noisy source state is straightforward to construct because the clean source state is available and can be directly noised at the current noise level. In contrast, the target-side state cannot be obtained in the same way because no corresponding clean target state is available. Instead, it must be constructed from the states already available during editing, without access to a target image. This makes target-side state construction a \emph{central challenge} in inversion-free flow-based editing.

A common solution is the \textbf{equal-displacement construction}~\citep{kulikov2025floweditinversionfreetextbasedediting}, which sets the displacement from the noisy source state to the target-side state \emph{equal to} that from the clean source state to the edited state, as illustrated in Fig.~\ref{fig:intro}(b). Subsequent works retain this construction while improving trajectory regularization, semantic decomposition, or source-feature integration~\citep{kim2025flowaligntrajectoryregularizedinversionfreeflowbased,yoon2025splitflowflowdecompositioninversionfree,jiang2025flowdcflowbaseddecouplingdecaycomplex,yang2025fiaeditfrequencyinteractiveattentionefficient}. Recent work has begun to explore alternatives. SNR-Edit~\citep{jiang2026snreditstructureawarenoiserectification} uses external structural information to modify the noise added to the clean source state, while DRFS~\citep{beaudouin2026deltarectifiedflowsampling} explores noise-level-dependent schedules that determine how much of the displacement between the clean source state and the edited state is incorporated into the target-side state. These works modify state construction choices but leave unexamined what the target-side state should approximate in inversion-free editing.

A closer examination reveals that the equal-displacement construction \emph{does not preserve how displacement changes under noising}. When the clean source state and the edited state are noised using the same noise level and sample, their displacement decreases as the noise level increases and vanishes at the pure-noise endpoint.\footnote{A formal derivation is provided in Appendix~\ref{app:method-analysis}.} The equal-displacement construction instead keeps the noisy displacement \emph{equal to} the clean displacement at every noise level, so the noisy source state and the target-side state remain separated even when dominated by noise and nearly indistinguishable. At high-noise steps, the target-prompt velocity is therefore evaluated at a target-side state too far from the noisy source state, which can produce overly aggressive updates and unwanted changes to edit-irrelevant content.

This suggests the target-side state should reflect how displacement changes under noising rather than preserve a fixed clean displacement. Since the clean target state is unavailable, we use the edited state as its evolving approximation and noise it with the same noise level and sample as the clean source state. Put simply, the equal-displacement construction keeps the noisy displacement \emph{equal to} the clean displacement, which is inconsistent with how displacement should behave under noising, whereas re-noising constructs both noisy states through shared noising, allowing their displacement to \emph{decrease naturally} as the noise level increases.

In this paper, we propose Re-Noising with Internal Dynamic Guidance for Image Editing (RIDGE), an inversion-free and training-free method built on this re-noising principle. RIDGE maintains the edited state as an evolving approximation to the unavailable clean target state. At each step, its \textbf{re-noising construction} noises this state using the same noise level and noise sample as the clean source state to obtain the target-side state, as illustrated in Fig.~\ref{fig:intro}(c).

However, during the early steps of editing, the edited state still contains limited target semantics since it is initialized from the clean source state. RIDGE therefore applies \textbf{internal dynamic guidance} to improve this weak approximation to the unavailable clean target state. RIDGE first applies the velocity-difference update to obtain a \emph{provisional edited state} and then obtains a \emph{clean target state prediction} under the target prompt as the guidance signal. Because this prediction can be uncertain at high noise levels, RIDGE applies guidance through a \emph{soft dynamic mask}, so the provisional edited state is guided toward the prediction only in regions requiring greater modification. The resulting edited state is then used by the re-noising construction, as illustrated in Fig.~\ref{fig:intro}(c). The guidance is \emph{internal} because both the prediction and the mask are produced by the editing model itself, without external segmentation or detection models. The mask is \emph{dynamic} as the edited state changes, allowing RIDGE to adapt which regions receive stronger guidance at different steps.

Extensive experiments on PIE-Bench Official and PIE-Bench++ under SD3 Medium and FLUX.1-dev show that RIDGE offers a favorable aggregate balance among the compared methods. Across these two benchmarks and two backbones, it combines strong source preservation with competitive target alignment and perceptual quality, covering standard local edits and diverse instructions. In summary, our contributions are as follows:
\begin{itemize}

\item We identify the under-examined problem of target-side state construction and show that the widely used equal-displacement construction fails to preserve how displacement changes under noising.

\item We formulate a \emph{re-noising construction} that maintains the edited state as an evolving approximation to the unavailable clean target state and noises it using the same noise level and noise sample as the clean source state.

\item We introduce \emph{internal dynamic guidance}, which uses a clean target state prediction and a soft dynamic mask to compensate for limited target semantics during early editing and strengthen changes in relevant regions without external tools.

\item On PIE-Bench Official and PIE-Bench++ under SD3 Medium and FLUX.1-dev, RIDGE achieves a favorable aggregate trade-off among source preservation, target alignment, global consistency, and perceptual quality while remaining inversion- and training-free.

\end{itemize}

\section{Related Work}

\paragraph{Diffusion-Based Image Editing.}
SDEdit~\citep{meng2022sdedit} edits real images through partial noising, Prompt-to-Prompt~\citep{hertz2023prompttoprompt} controls cross-attention, and PnP Inversion~\citep{ju2024pnpinversion} improves latent inversion for diffusion-based editing. InfEdit~\citep{xu2024infedit} avoids explicit inversion through consistency sampling and unified attention control. Recent flow-based methods include Stable Flow~\citep{avrahami2025stableflow}, RF-Inversion~\citep{rout2024semanticimageinversionediting}, RF-Solver/RF-Edit~\citep{wang2025tamingrectifiedflowinversion}, and Unveil Inversion and Invariance~\citep{xu2025unveilinversioninvarianceflow}, which adapt these paradigms through specialized solvers and feature injection.

\paragraph{Inversion-Free Flow-Based Editing.}
FlowEdit~\citep{kulikov2025floweditinversionfreetextbasedediting} introduced an inversion-free formulation that edits an image using the difference between the velocities conditioned on the source and target prompts. To obtain these velocities, it constructs the target-side state using an \emph{equal-displacement construction}, which sets the noisy displacement equal to the clean displacement across noise levels. Subsequent work improves different parts of this editing process. FlowAlign~\citep{kim2025flowaligntrajectoryregularizedinversionfreeflowbased} regularizes the editing path to better preserve the source image. SplitFlow~\citep{yoon2025splitflowflowdecompositioninversionfree} decomposes the target prompt into multiple semantic flows to better handle edits involving several concepts. FlowDC~\citep{jiang2025flowdcflowbaseddecouplingdecaycomplex} separates multiple editing effects and suppresses velocity components less relevant to the requested changes. FIA-Edit~\citep{yang2025fiaeditfrequencyinteractiveattentionefficient} incorporates source information through frequency-interactive attention and feature injection. These methods improve the editing update, trajectory, or source preservation, while largely retaining FlowEdit's equal-displacement construction. This motivates a closer look at target-side state construction itself.

\paragraph{State Construction in Inversion-Free Flow Editing.}
Other works modify the states at which source- and target-prompt velocities are evaluated. Beyond equal-displacement constructions as in FlowEdit~\citep{kulikov2025floweditinversionfreetextbasedediting}, SNR-Edit~\citep{jiang2026snreditstructureawarenoiserectification} rectifies sampled noise using structural information from external tools; DRFS~\citep{beaudouin2026deltarectifiedflowsampling} introduces noise-level-dependent shifts that vary the target-side construction; and FlowCycle~\citep{wang2026targetawareimageeditingcycleconsistent} optimizes corruption noise for a target-aware intermediate state under cycle-consistency constraints. These approaches alter source initialization, displacement scheduling, or the corruption process. RIDGE instead focuses on what the target-side state should approximate during direct source-to-target transport, and constructs it from an evolving approximation to the unavailable clean target state.

\section{Methodology}

\begin{figure}[t]
    \centering
    \includegraphics[width=\linewidth]{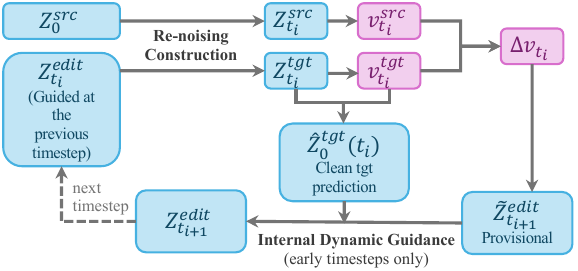}
    \caption{Overview of one RIDGE iteration from timestep \(t_i\) to \(t_{i+1}\). The soft dynamic mask applied during guidance is omitted for clarity.}
    \label{fig:pipeline}
\end{figure}

\subsection{Preliminaries: Rectified Flow}
\label{sec:preliminaries}

Rectified flow learns a velocity field from linear interpolations between clean data and Gaussian noise~\citep{liu2022flowstraightfastlearning,lipman2023flowmatchinggenerativemodeling}. Let \(X\sim\pi_0\) be a clean latent and \(\eta\sim\pi_1=\mathcal{N}(0,I)\). Their interpolation is
\begin{equation}
Z_t=(1-t)X+t\eta,
\qquad t\in[0,1],
\label{eq:pre_rf_path}
\end{equation}
where \(t=0\) corresponds to clean data and \(t=1\) to Gaussian noise. The velocity along this path is
\begin{equation}
u_t
=
\frac{dZ_t}{dt}
=
\eta-X.
\label{eq:pre_rf_velocity}
\end{equation}
A text-conditioned predictor \(v_\theta(Z_t,c,t)\) estimates this field.

At inference, generation starts from \(Z_1=\eta\) and solves
\begin{equation}
\frac{dZ_t}{dt}
=
v_\theta(Z_t,c,t)
\label{eq:pre_rf_ode}
\end{equation}
from \(t=1\) to \(t=0\) under the discrete schedule
\begin{equation}
1=t_0>t_1>\cdots>t_N=0.
\label{eq:pre_rf_schedule}
\end{equation}
An Euler step then updates \(Z_{t_i}\) for \(i=0,\ldots,N-1\):
\begin{equation}
Z_{t_{i+1}}
=
Z_{t_i}
+
(t_{i+1}-t_i)
v_\theta
\left(
Z_{t_i},
c,
t_i
\right).
\label{eq:pre_rf_euler}
\end{equation}

\subsection{RIDGE Overview}

We present \emph{Re-Noising with Internal Dynamic Guidance for Image Editing (RIDGE)}, an inversion-free and training-free latent editing method. Given a clean source state \(Z_0^{\mathrm{src}}\), a source prompt \(c^{\mathrm{src}}\), and a target prompt \(c^{\mathrm{tgt}}\), RIDGE treats the edited state as an evolving approximation to the unavailable clean target state. Its \emph{re-noising construction} noises this approximation at the current noise level to obtain the target-side state. Because the edited state is initialized from the clean source state and initially contains limited target semantics, \emph{internal dynamic guidance} refines this weak approximation during the early high-noise steps.

Figure~\ref{fig:pipeline} summarizes one RIDGE iteration from \(t_i\) to \(t_{i+1}\). \textbf{1) Re-noising construction.} The edited state \(Z_{t_i}^{\mathrm{edit}}\) from the preceding iteration and the clean source state \(Z_0^{\mathrm{src}}\) are noised with the same noise level and noise sample, yielding the target-side state \(Z_{t_i}^{\mathrm{tgt}}\) and noisy source state \(Z_{t_i}^{\mathrm{src}}\), respectively. The difference \(\Delta v_{t_i}\) between the target- and source-prompt velocities evaluated at these states updates \(Z_{t_i}^{\mathrm{edit}}\) to the provisional edited state \(\widetilde Z_{t_{i+1}}^{\mathrm{edit}}\). \textbf{2) Internal dynamic guidance.} During early high-noise steps, the target branch additionally predicts the clean target state \(\hat Z_0^{\mathrm{tgt}}(t_i)\). A soft dynamic mask \(m_{t_i}\), derived from this prediction and the clean source state, restricts guidance to regions requiring greater modification. The guidance update yields the next edited state \(Z_{t_{i+1}}^{\mathrm{edit}}\), which is used as the clean-state approximation in the subsequent re-noising step. Repeating this process along the reverse-time schedule produces the final clean edited result \(Z_0^{\mathrm{edit}}\).

\begin{figure}[t]
    \centering
    \includegraphics[width=\linewidth]{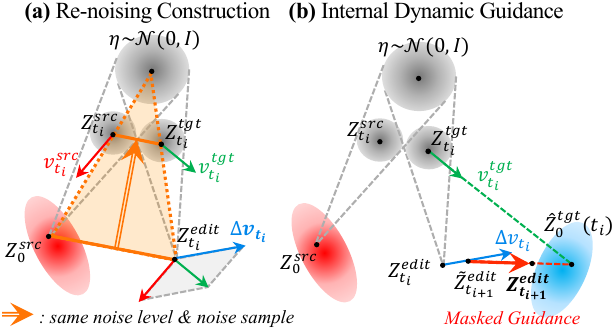}
    \caption{(a) The re-noising construction forms the noisy source and target-side states used in the velocity-difference update. (b) Internal dynamic guidance applies a masked correction toward the clean target state prediction during early high-noise steps.}
    \label{fig:ridge_geometry}
\end{figure}

\subsection{Re-Noising Construction}

\paragraph{Editing window.}
RIDGE applies velocity-difference editing only for \(t_{\min}<t_i\le t_{\max}\), initializing \(Z_{t_{\max}}^{\mathrm{edit}}=Z_0^{\mathrm{src}}\) at the upper boundary. The edited state then follows recursively from the updates below. Upon reaching \(t_{\min}\), RIDGE stops the velocity-difference updates and completes the remaining reverse-time steps to \(t=0\) under the target prompt.

\paragraph{Shared noising.}
At \(t_i\), RIDGE samples \(\eta_{t_i}\sim\mathcal{N}(0,I)\) and noises the clean source and current edited states with this shared noise sample:
\begin{equation}
\begin{aligned}
Z_{t_i}^{\mathrm{src}}
&=
(1-t_i)Z_0^{\mathrm{src}}+t_i\eta_{t_i}, \\
Z_{t_i}^{\mathrm{tgt}}
&=
(1-t_i)Z_{t_i}^{\mathrm{edit}}+t_i\eta_{t_i}.
\end{aligned}
\label{eq:ridge_renoising}
\end{equation}

Their difference satisfies
\begin{equation}
Z_{t_i}^{\mathrm{tgt}}-Z_{t_i}^{\mathrm{src}}
=
(1-t_i)
\left(
Z_{t_i}^{\mathrm{edit}}-Z_0^{\mathrm{src}}
\right).
\label{eq:ridge_noisy_displacement}
\end{equation}
The factor \(1-t_i\) is fixed by applying the rectified-flow noising path to both clean-state estimates with shared noise, rather than being selected as an independent displacement schedule. The resulting displacement vanishes at the pure-noise endpoint instead of remaining constant across noise levels, as represented by the orange triangular region in Fig.~\ref{fig:ridge_geometry}(a).

\paragraph{Velocity-difference update.}
The target-minus-source velocity difference is
\begin{equation}
\Delta v_{t_i}
=
v_\theta
\left(
Z_{t_i}^{\mathrm{tgt}},
c^{\mathrm{tgt}},
t_i
\right)
-
v_\theta
\left(
Z_{t_i}^{\mathrm{src}},
c^{\mathrm{src}},
t_i
\right).
\label{eq:ridge_velocity_difference}
\end{equation}
The green and red arrows in Fig.~\ref{fig:ridge_geometry}(a) represent the two velocity terms, and the blue arrow represents \(\Delta v_{t_i}\). The provisional edited state is then
\begin{equation}
\widetilde Z_{t_{i+1}}^{\mathrm{edit}}
=
Z_{t_i}^{\mathrm{edit}}
+
(t_{i+1}-t_i)\Delta v_{t_i}.
\label{eq:ridge_velocity_update}
\end{equation}
Figure~\ref{fig:ridge_geometry}(b) illustrates this update from \(Z_{t_i}^{\mathrm{edit}}\) to \(\widetilde Z_{t_{i+1}}^{\mathrm{edit}}\).

\subsection{Internal Dynamic Guidance}

During the early high-noise steps, the edited state contains limited target semantics. RIDGE therefore applies internal dynamic guidance over \(t_{\mathrm{end}}\le t_i\le t_{\mathrm{start}}\) within the editing window to better approximate the clean target state.

\paragraph{Clean target state prediction.}
The target branch provides the clean-state prediction:
\begin{equation}
\hat Z_0^{\mathrm{tgt}}(t_i)
=
Z_{t_i}^{\mathrm{tgt}}
-
t_i
v_\theta
\left(
Z_{t_i}^{\mathrm{tgt}},
c^{\mathrm{tgt}},
t_i
\right).
\label{eq:ridge_x0_prediction}
\end{equation}
The prediction computed at \(t_i\) guides the provisional edited state \(\widetilde Z_{t_{i+1}}^{\mathrm{edit}}\), yielding a \emph{one-step-delayed approximation}. See Appendix~\ref{app:method-analysis} for the rationale.

\paragraph{Soft dynamic mask.}
Predictions at high noise levels can exhibit \emph{unstable} spatial change patterns. Applying guidance uniformly would propagate this uncertainty into edit-irrelevant regions. RIDGE therefore employs a soft dynamic mask that emphasizes locations with larger predicted changes and attenuates the remainder.

RIDGE computes a spatial difference map by averaging absolute residuals across channels:
\begin{equation}
D_{t_i}
=
\frac{1}{C}
\sum_{c=1}^{C}
\left|
\hat Z_{0,c}^{\mathrm{tgt}}(t_i)
-
Z_{0,c}^{\mathrm{src}}
\right|,
\label{eq:ridge_diff_map}
\end{equation}
where \(C\) is the number of latent channels and subscript \(c\) indexes the channel. The map \(D_{t_i}\) preserves the spatial dimensions and quantifies the predicted change at each location.

Rather than using a fixed threshold across images and timesteps, RIDGE identifies relatively large changes from the spatial distribution of \(D_{t_i}\). It sets the adaptive threshold to the \(q\)-quantile of \(D_{t_i}\):
\begin{equation}
\phi_{t_i}
=
\operatorname{Quantile}_{q}
\left(
D_{t_i}
\right).
\label{eq:ridge_mask_threshold}
\end{equation}
The difference map is then converted into a soft mask:
\begin{equation}
m_{t_i}
=
\sigma\!\left(
\frac{D_{t_i}-\phi_{t_i}}{\tau}
\right),
\label{eq:ridge_mask_sigmoid}
\end{equation}
where \(\sigma(\cdot)\) is applied element-wise and \(\tau>0\) controls smoothness around the threshold. The resulting mask assigns larger weights to locations with larger predicted changes while attenuating weaker differences. Its construction uses only quantities available during editing and requires no external segmentation model.

\paragraph{Guided update.}
Within \(t_{\mathrm{end}}\le t_i\le t_{\mathrm{start}}\), RIDGE refines the provisional edited state toward the clean target state prediction:
\begin{equation}
Z_{t_{i+1}}^{\mathrm{edit}}
=
\widetilde Z_{t_{i+1}}^{\mathrm{edit}}
+
\gamma_{t_i}m_{t_i}\odot
\left(
\hat Z_0^{\mathrm{tgt}}(t_i)
-
\widetilde Z_{t_{i+1}}^{\mathrm{edit}}
\right).
\label{eq:ridge_guidance_update}
\end{equation}
The residual in parentheses defines the \emph{guidance direction} from the provisional edited state toward the clean target state prediction, while \(m_{t_i}\) assigns larger weights to regions estimated as more relevant to the edit. The red arrow in Fig.~\ref{fig:ridge_geometry}(b) illustrates this spatially weighted correction.

The scalar \(\gamma_{t_i}\) controls the guidance strength:
\begin{equation}
\gamma_{t_i}
=
\frac{\lambda_{\mathrm{guide}}}
{1-\beta_{\mathrm{guide}}t_i},
\label{eq:ridge_guidance_strength}
\end{equation}
where \(\lambda_{\mathrm{guide}}\) sets the base strength and \(\beta_{\mathrm{guide}}\) controls its noise-level dependence. Guidance is therefore stronger at larger \(t_i\) and decreases along the reverse-time trajectory.

Outside the guidance window, after the edited state has accumulated stronger target semantics, the provisional velocity-difference update is accepted without additional guidance, \ie, \(Z_{t_{i+1}}^{\mathrm{edit}}=\widetilde Z_{t_{i+1}}^{\mathrm{edit}}\). At \(t_{i+1}\), the resulting edited state enters the next iteration, where shared re-noising constructs the new noisy source and target-side states, and the same update cycle repeats along the reverse-time schedule.

\subsection{RIDGE Algorithm}

Alg.~\ref{alg:ridge} presents a simplified RIDGE procedure; the full algorithm is provided in Appendix~\ref{app:full-algorithm}. Masked guidance is applied within the specified window. For compactness, \(\mathcal{M}(\cdot,\cdot)\) denotes the mask construction in Eqs.~\ref{eq:ridge_diff_map}--\ref{eq:ridge_mask_sigmoid}, and \(\gamma_{t_i}\) follows Eq.~\ref{eq:ridge_guidance_strength}.

\begin{algorithm}[t]
\caption{Simplified algorithm for RIDGE}
\label{alg:ridge}
\begin{algorithmic}[1]
\Require Clean source state \(Z_0^{\mathrm{src}}\), prompts \(c^{\mathrm{src}}\) and \(c^{\mathrm{tgt}}\), reverse-time schedule \(\{t_i\}\), editing window, guidance window, guidance schedule \(\gamma_{t_i}\) defined by Eq.~\ref{eq:ridge_guidance_strength}, velocity predictor \(v_\theta\).
\Ensure Clean edited state \(Z_0^{\mathrm{edit}}\).

\State Initialize \(Z_{t_{\max}}^{\mathrm{edit}}\gets Z_0^{\mathrm{src}}\).

\For{each editing timestep \(t_i\)}
    \State Sample \(\eta_{t_i}\sim\mathcal{N}(0,I)\).
    \State \(Z_{t_i}^{\mathrm{src}}\gets (1-t_i)Z_0^{\mathrm{src}}+t_i\eta_{t_i}\).
    \State \(Z_{t_i}^{\mathrm{tgt}}\gets (1-t_i)Z_{t_i}^{\mathrm{edit}}+t_i\eta_{t_i}\).

    \State \(\Delta v_{t_i}\gets
    v_\theta(Z_{t_i}^{\mathrm{tgt}},c^{\mathrm{tgt}},t_i)
    -
    v_\theta(Z_{t_i}^{\mathrm{src}},c^{\mathrm{src}},t_i)\).

    \State \(\widetilde Z_{t_{i+1}}^{\mathrm{edit}}\gets
    Z_{t_i}^{\mathrm{edit}}
    +
    (t_{i+1}-t_i)\Delta v_{t_i}\).

    \If{\(t_i\) lies within the guidance window}
        \State \(\hat Z_0^{\mathrm{tgt}}(t_i)\gets
        Z_{t_i}^{\mathrm{tgt}}
        -
        t_i v_\theta(Z_{t_i}^{\mathrm{tgt}},c^{\mathrm{tgt}},t_i)\).

        \State \(m_{t_i}\gets
        \mathcal{M}\!\left(\hat Z_0^{\mathrm{tgt}}(t_i),Z_0^{\mathrm{src}}\right)\).

        \State \(Z_{t_{i+1}}^{\mathrm{edit}}\gets
        \widetilde Z_{t_{i+1}}^{\mathrm{edit}}
        +
        \gamma_{t_i}m_{t_i}\odot
        \left(
        \hat Z_0^{\mathrm{tgt}}(t_i)
        -
        \widetilde Z_{t_{i+1}}^{\mathrm{edit}}
        \right)\).
    \Else
        \State \(Z_{t_{i+1}}^{\mathrm{edit}}\gets
        \widetilde Z_{t_{i+1}}^{\mathrm{edit}}\).
    \EndIf
\EndFor

\State Complete the remaining target-prompt flow steps.
\State \Return \(Z_0^{\mathrm{edit}}\).
\end{algorithmic}
\end{algorithm}

\section{Experiments}

\begin{table*}[t]
\centering
{\small
\setlength{\tabcolsep}{1.5pt}
\caption{Quantitative comparison on PIE-Bench Official and PIE-Bench++.
RF-Inv. denotes RF-Inversion. Best, second-best, and third-best values (per dataset) are shown in boldface, boxed, and underlined, respectively.}
\label{tab:pie_bench_selected_metrics}
\begin{tabular}{@{}cllccccccc@{}}
\toprule
\multirow{2}{*}{Dataset} & \multirow{2}{*}{Type} & \multirow{2}{*}{Method}
& \multicolumn{4}{c}{Source Preservation}
& \multicolumn{1}{c}{Target Alignment}
& \multicolumn{1}{c}{Aesthetic}
& \multirow{2}{*}{Avg Sc.$\uparrow$} \\
\cmidrule(lr){4-7}\cmidrule(lr){8-8}\cmidrule(lr){9-9}
& & & Struct.$\downarrow$ & PSNR$_{\rm u}\uparrow$ & LPIPS$_{\rm u}\downarrow$ & DINO$\downarrow$ & CLIP$_{\rm tgt}\uparrow$ & Aes.$\uparrow$ & \\
\midrule
\multirow{6}{*}{PIE-Bench} & Inv-B & RF-Inv.
& 0.047 & 19.752 & 0.193 & 0.434 & 26.811 & \third{5.799} & 0.204 \\
& Inv-B & UniEdit
& \third{0.021} & \second{25.260} & 0.094 & \second{0.328} & 26.481 & 5.725 & 0.620 \\
& Inv-F & FlowEdit
& 0.039 & 20.250 & 0.140 & 0.420 & \best{27.688} & \best{5.807} & 0.476 \\
& Inv-F & FlowAlign
& \best{0.016} & \best{26.267} & \best{0.057} & \third{0.343} & 26.692 & 5.650 & \third{0.648} \\
& Inv-F & DRFS
& 0.023 & 23.400 & \third{0.089} & 0.351 & \second{27.535} & 5.777 & \second{0.742} \\
\oursrow
& Inv-F & \textbf{RIDGE}
& \second{0.021} & \third{23.570} & \second{0.085} & \best{0.306} & \third{27.137} & \second{5.804} & \best{0.794} \\
\midrule
\multirow{6}{*}{PIE-Bench++} & Inv-B & RF-Inv.
& 0.048 & 20.274 & 0.177 & 0.498 & 26.947 & \third{5.779} & 0.187 \\
& Inv-B & UniEdit
& 0.026 & \second{25.540} & 0.095 & \third{0.428} & 26.482 & 5.689 & 0.541 \\
& Inv-F & FlowEdit
& 0.042 & 20.635 & 0.132 & 0.491 & \best{28.121} & \best{5.806} & 0.456 \\
& Inv-F & FlowAlign
& \second{0.023} & \best{26.382} & \best{0.059} & 0.474 & 26.941 & 5.641 & \third{0.573} \\
& Inv-F & DRFS
& \third{0.025} & 23.889 & \third{0.082} & \second{0.428} & \second{27.891} & 5.776 & \second{0.751} \\
\oursrow
& Inv-F & \textbf{RIDGE}
& \best{0.022} & \third{24.007} & \second{0.081} & \best{0.371} & \third{27.226} & \second{5.791} & \best{0.798} \\
\bottomrule
\end{tabular}
}
\end{table*}

\subsection{Experimental Setup}

\paragraph{Benchmarks.}
We evaluate RIDGE on two text-based image editing benchmarks: PIE-Bench Official~\citep{ju2024pnpinversion} and PIE-Bench++~\citep{huang2024paralleledits}. PIE-Bench++ extends PIE-Bench Official to multifaceted editing scenarios involving multiple objects and attributes.

\paragraph{Baselines.}
Under Stable Diffusion 3 Medium (SD3 Medium)~\citep{esser2024scalingrectifiedflowtransformers}, we compare RIDGE with the inversion-based (Inv-B) methods RF-Inversion (RF-Inv.)~\citep{rout2024semanticimageinversionediting} and UniEdit~\citep{jiao2026unieditflowunleashinginversionediting}, and the inversion-free (Inv-F) methods FlowEdit~\citep{kulikov2025floweditinversionfreetextbasedediting}, FlowAlign~\citep{kim2025flowaligntrajectoryregularizedinversionfreeflowbased}, and DRFS~\citep{beaudouin2026deltarectifiedflowsampling}. Under FLUX.1-dev, we evaluate the compatible subset: RIDGE, RF-Inversion, UniEdit, and FlowEdit.


\paragraph{Implementation details.}
Images are cropped to multiples of \(16\) at native resolution. We evaluate all 700 PIE-Bench Official edits and use the same protocol on PIE-Bench++. Baselines use their public implementations and recommended configurations when available; FLUX uses the compatible subset above. For a controlled SD3 comparison, RIDGE matches FlowEdit's editing window and classifier-free guidance settings~\citep{ho2022classifierfreediffusionguidance}: \(T_{\mathrm{steps}}=50\), \(n_{\max}=36\), \(n_{\min}=5\), and source/target CFG scales of \(3.5\) and \(13.5\).
RIDGE samples fresh Gaussian noise at each step (\(n_{\mathrm{avg}}=1\)) and applies internal dynamic guidance over \(n\in[30,36]\), with \(\lambda_{\mathrm{guide}}=0.012\), \(\beta_{\mathrm{guide}}=0.02\), mask quantile \(q=0.7\), and temperature \(\tau=0.2\).
Per-method configurations and additional implementation details are provided in Appendix~\ref{app:exp-details}.

\paragraph{Evaluation metrics.}
We evaluate source preservation, target alignment, and aesthetic quality. For source preservation, StructDist~\citep{tumanyan2022splicing,ju2024pnpinversion} measures full-image spatial structure through DINO key self-similarity~\citep{caron2021emergingpropertiesselfsupervisedvision}, while PSNR$_{\rm u}$ and LPIPS$_{\rm u}$~\citep{zhang2018unreasonableeffectivenessdeepfeatures} respectively measure pixel-level and perceptual fidelity in unedited regions. DINOv2 CLS cosine distance (DINO)~\citep{oquab2024dinov2learningrobustvisual} measures global visual consistency, following prior DINO-based input--edit evaluations~\citep{sheynin2024emuedit}. CLIP$_{\rm tgt}$~\citep{radford2021learningtransferablevisualmodels} measures target-prompt alignment, and Aes. measures predicted aesthetic quality. In the tables, Struct., Aes., and Avg Sc. denote StructDist, the LAION aesthetic score, and Avg Score, respectively; the subscripts \({\rm u}\) and \({\rm tgt}\) indicate the unedited region and target prompt. Arrows indicate whether higher or lower values are better. Metric implementation details are provided in Appendix~\ref{app:exp-details}.

\paragraph{Average score.}
Since the metrics have different scales and optimization directions, we additionally report \emph{Avg Score} as a compact aggregate overview of the six metrics. Within each dataset, we min--max normalize every metric across the compared methods, invert lower-is-better metrics, and take their equal-weighted mean.

\paragraph{Additional evaluations.}
The main table reports SD3 results on PIE-Bench Official and PIE-Bench++; Appendix~\ref{app:add-quant} reports FLUX.1-dev results on both benchmarks, and Appendix~\ref{app:user-study} reports the human preference study.

\begin{table}[t]
\centering
{\small
\setlength{\tabcolsep}{2pt}
\caption{Ablation of internal dynamic guidance and its soft dynamic mask on PIE-Bench Official. ``w/o guidance'' removes the guided update, while ``w/o mask'' applies guidance globally.}
\label{tab:ablation_mask}
\begin{tabular}{@{}lccccc@{}}
\toprule
Method
& Struct.$\downarrow$
& PSNR$_{\rm u}\uparrow$
& LPIPS$_{\rm u}\downarrow$
& DINO$\downarrow$
& CLIP$_{\rm tgt}\uparrow$ \\
\midrule
RIDGE
& 0.021
& 23.570
& 0.085
& 0.306
& 27.137 \\
w/o guidance
& 0.015
& 25.260
& 0.069
& 0.267
& 26.780 \\
w/o mask
& 0.024
& 23.071
& 0.096
& 0.323
& 27.211 \\
\bottomrule
\end{tabular}
}
\end{table}

\begin{figure}[t]
    \centering
    \includegraphics[width=\linewidth]{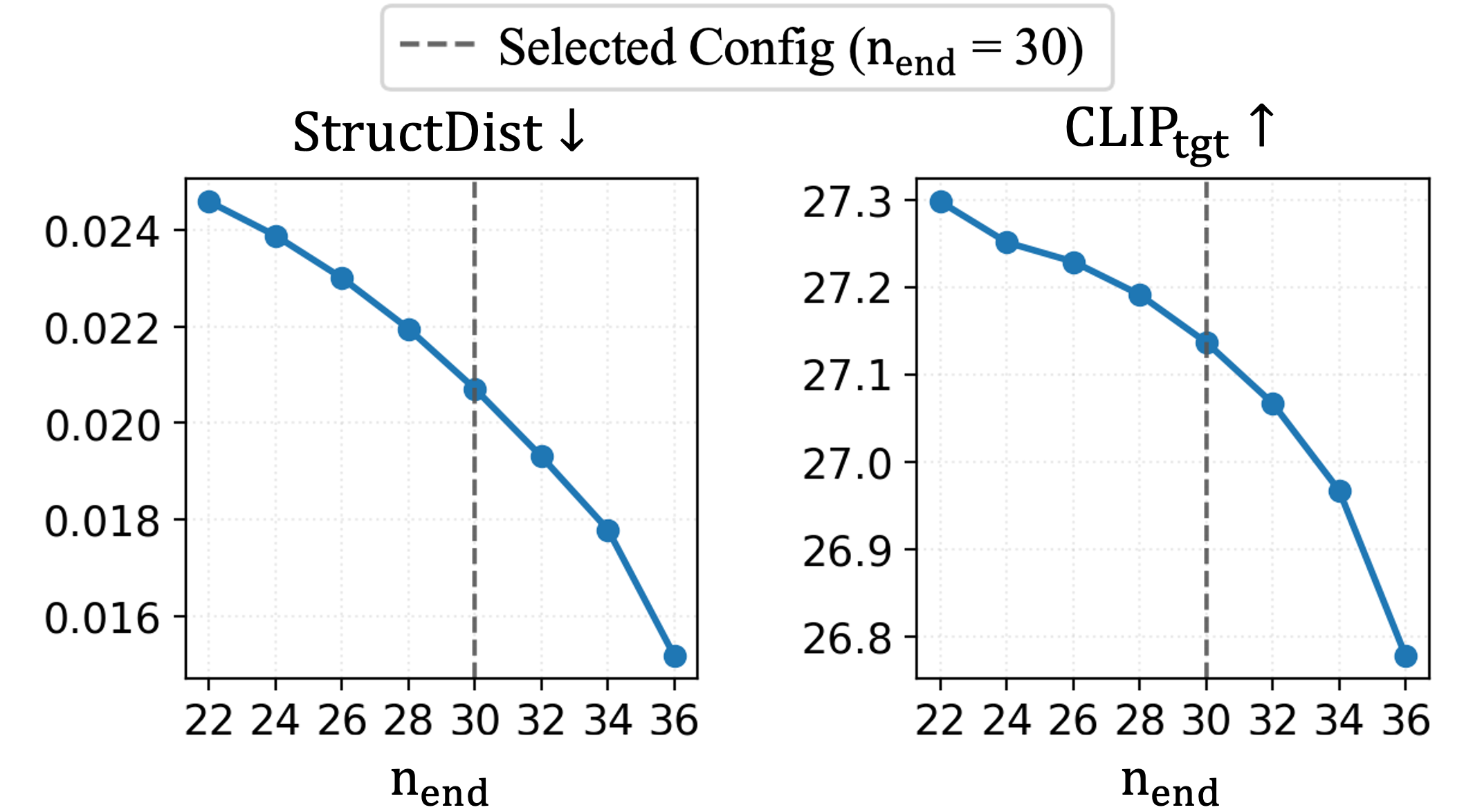}
    \caption{Guidance-window sweep on PIE-Bench Official. The dashed line marks the default setting (\(n_{\mathrm{end}}=30\)).}
    \label{fig:ablation_guidance_window}
\end{figure}

\subsection{Main Results}

Table~\ref{tab:pie_bench_selected_metrics} reports results on PIE-Bench Official and PIE-Bench++. RIDGE achieves strong source preservation on both benchmarks, while maintaining competitive target alignment and aesthetic quality. In particular, it obtains the lowest DINO distance on both datasets and the best StructDist on PIE-Bench++, indicating better global visual and structural preservation. Under the equal-weighted normalization defined above, RIDGE ranks first in aggregate Avg Score on both benchmarks, indicating a favorable trade-off between editing strength and source preservation.

\begin{figure*}[t]
    \centering
    \includegraphics[width=\textwidth]{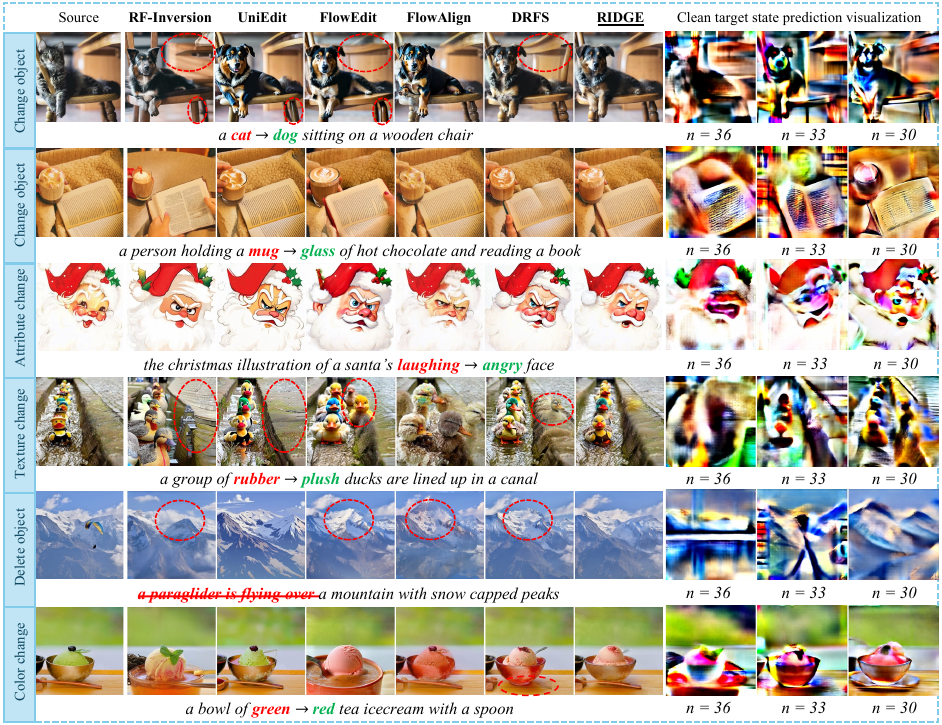}
    \caption{Qualitative comparisons on representative editing examples across different edit types. Columns show the source image, results from baseline methods and RIDGE, and RIDGE's clean target state predictions at \(n=36,33,30\). Red dashed circles highlight representative baseline failure modes.}
    \label{fig:qual_comparisons}
\end{figure*}

\subsection{Qualitative Results}

Figure~\ref{fig:qual_comparisons} shows representative qualitative comparisons across different edit types, including object change, attribute change, texture change, object deletion, and color change. Existing methods often exhibit different edit--preserve imbalances: some produce strong target-directed changes but alter edit-irrelevant regions, while others better preserve the source image but under-edit challenging semantic or texture changes. In contrast, RIDGE tends to preserve more edit-irrelevant structure such as background layout, pose, and surrounding objects, while still moving the image sufficiently toward the target prompt.

The rightmost columns visualize RIDGE's clean target state predictions at selected guidance steps, which generally become more stable and aligned with the intended edit at lower noise levels. More prediction visualizations and timestep-wise diagnostics are in Appendix~\ref{app:pred-analysis}.

\subsection{Ablation}

\paragraph{Masked guidance.}
Internal dynamic guidance is a central component of RIDGE. We disentangle the effects of internal dynamic guidance and its spatial mask in Table~\ref{tab:ablation_mask}. The ``w/o guidance'' variant removes the guided update entirely, leaving the re-noising-based velocity update unchanged. It achieves stronger source preservation but lower CLIP$_{\rm tgt}$, indicating that re-noising without guidance is conservative and that early guidance strengthens target semantics. The ``w/o mask'' variant retains guidance but sets \(m_{t_i}=1\) everywhere. Applying guidance globally yields the highest CLIP$_{\rm tgt}$ but degrades all four preservation metrics relative to full RIDGE. Masked guidance therefore provides the intended division of roles: guidance increases editing strength, while the dynamic mask concentrates the correction on regions predicted to require modification, recovering source preservation with only a \(0.074\) decrease in CLIP$_{\rm tgt}$ relative to global guidance.

\begin{figure}[t]
    \centering
    \includegraphics[width=\linewidth]{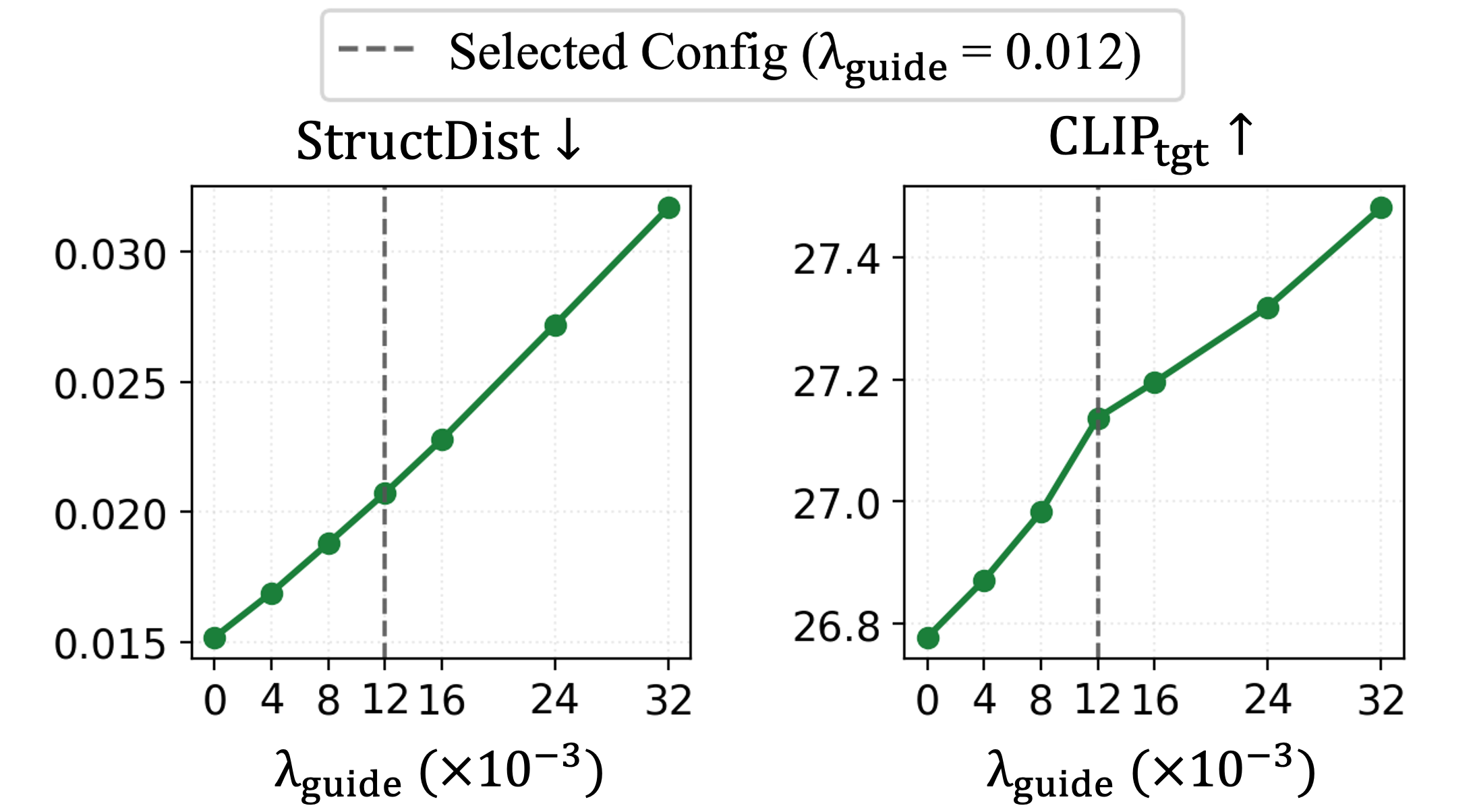}
    \caption{Guidance-strength sweep on PIE-Bench Official. The dashed line marks the default setting (\(\lambda_{\mathrm{guide}}=0.012\)).}
    \label{fig:ablation_lambda}
\end{figure}

\paragraph{Guidance window.}
To study guidance duration, we fix \(n_{\mathrm{start}}=36\) and sweep \(n_{\mathrm{end}}\) from 34 down to 22 on PIE-Bench Official, keeping all other hyperparameters fixed. Fig.~\ref{fig:ablation_guidance_window} shows that longer guidance (smaller \(n_{\mathrm{end}}\)) improves CLIP$_{\rm tgt}$ but increases StructDist. We select \(n_{\mathrm{end}}=30\) as a balanced operating point; additional metrics are provided in Appendix~\ref{app:add-ablation}.

\paragraph{Guidance strength.}
To study guidance strength, we sweep \(\lambda_{\mathrm{guide}}\in\{0,0.004,0.008,0.012,0.016,0.024,0.032\}\) in Eq.~\ref{eq:ridge_guidance_strength} on PIE-Bench Official, keeping all other hyperparameters fixed. Fig.~\ref{fig:ablation_lambda} shows that larger \(\lambda_{\mathrm{guide}}\) improves CLIP$_{\rm tgt}$ but increases StructDist. We select \(\lambda_{\mathrm{guide}}=0.012\) as a balanced operating point; additional metrics are provided in Appendix~\ref{app:add-ablation}.

\section{Conclusion and Limitations}

We presented RIDGE, an inversion-free and training-free method that reconsiders target-side state construction in flow-based image editing. RIDGE treats the evolving edited state as an approximation to the unavailable clean target state and re-noises it with the clean source state, allowing their displacement to contract naturally under noising. Internal dynamic guidance strengthens early target semantics while focusing correction on regions requiring modification. Experiments on two benchmarks and two backbones show a strong balance among source preservation, target alignment, and perceptual quality. Like many inversion-free approaches, RIDGE may remain conservative for edits involving major changes in object identity, geometry, or scene composition. Nevertheless, RIDGE interprets the unknown target-side state by approximating the unavailable clean target state and then noising it, rather than equating noisy and clean displacements. We hope this work highlights target-side state construction as a central design choice in inversion-free flow editing.

\FloatBarrier
\clearpage
\bibliographystyle{abbrvnat}
\bibliography{references}

\clearpage
\onecolumn
\appendix
\setcounter{secnumdepth}{2}
\titleformat{\section}{\Large\bfseries}{\thesection}{0.75em}{}
\titleformat{\subsection}{\large\bfseries}{\thesubsection}{0.65em}{}
\titlespacing*{\section}{0pt}{1.8ex plus 0.4ex minus 0.2ex}{0.8ex}
\titlespacing*{\subsection}{0pt}{1.4ex plus 0.3ex minus 0.2ex}{0.5ex}

\begin{center}
{\LARGE\bfseries Appendix\par}
\end{center}
\vspace{0.8em}

\section{Method Analysis}
\label{app:method-analysis}
\subsection{Displacement Contraction under Shared Noising}
\label{app:noisy_displacement}

We derive the displacement induced by noising two clean states with the same noise level and noise sample. At an editing timestep \(t_i\), let \(Z_0^{\mathrm{src}}\) denote the clean source state and \(Z_{t_i}^{\mathrm{edit}}\) denote the edited state at the current editing step.

Under rectified-flow interpolation, applying shared noising with \(\eta_{t_i}\) gives
\begin{align}
Z_{t_i}^{\mathrm{src}}
&=
(1-t_i)Z_0^{\mathrm{src}}
+
t_i\eta_{t_i},
\label{eq:app_noisy_src}
\\
Z_{t_i}^{\mathrm{tgt}}
&=
(1-t_i)Z_{t_i}^{\mathrm{edit}}
+
t_i\eta_{t_i}.
\label{eq:app_noisy_tgt}
\end{align}
Subtracting Eq.~\eqref{eq:app_noisy_src} from Eq.~\eqref{eq:app_noisy_tgt} cancels the shared noise term:
\begin{align}
Z_{t_i}^{\mathrm{tgt}}
-
Z_{t_i}^{\mathrm{src}}
&=
(1-t_i)
\left(
Z_{t_i}^{\mathrm{edit}}
-
Z_0^{\mathrm{src}}
\right).
\label{eq:app_shared_noising_displacement}
\end{align}
Since \(t_i\in[0,1]\), taking the Euclidean norm yields
\begin{equation}
\|Z_{t_i}^{\mathrm{tgt}}-Z_{t_i}^{\mathrm{src}}\|_2
=
(1-t_i)
\|Z_{t_i}^{\mathrm{edit}}-Z_0^{\mathrm{src}}\|_2.
\label{eq:app_shared_noising_norm}
\end{equation}
Thus, for a fixed pair of clean states at timestep \(t_i\), shared noising scales their displacement by \(1-t_i\). As the noise level increases, this factor decreases from \(1\) to \(0\). At the pure-noise endpoint \(t_i=1\), both noised states collapse to the same noise sample:
\begin{equation}
Z_1^{\mathrm{src}}
=
Z_1^{\mathrm{tgt}}
=
\eta_1,
\end{equation}
and their displacement vanishes.

In contrast, the equal-displacement construction sets
\begin{equation}
Z_{t_i}^{\mathrm{tgt}}
-
Z_{t_i}^{\mathrm{src}}
=
Z_{t_i}^{\mathrm{edit}}
-
Z_0^{\mathrm{src}},
\end{equation}
so the noisy displacement is kept equal to the clean-state displacement at all noise levels. Unless the two clean states are identical, this displacement does not contract as \(t_i\) increases and would remain nonzero at the pure-noise endpoint. Therefore, equal displacement does not match the contraction behavior induced by noising.

\subsection{One-Step Delayed Guidance Approximation}

\paragraph{Same-timestep circular dependency.}
A seemingly more direct alternative is to first predict a clean target state at timestep \(t_{i+1}\) and use it to guide the edited state at the same timestep. However, this creates the following circular dependency:
\begin{equation}
\hat Z_0^{\mathrm{tgt}}(t_{i+1})
\;\longleftarrow\;
Z_{t_{i+1}}^{\mathrm{tgt}}
\;\longleftarrow\;
Z_{t_{i+1}}^{\mathrm{edit}}
\;\longleftarrow\;
\hat Z_0^{\mathrm{tgt}}(t_{i+1}),
\label{eq:app_guidance_circular_dependency}
\end{equation}
where each arrow denotes a dependency. Computing \(\hat Z_0^{\mathrm{tgt}}(t_{i+1})\) requires the target-side state \(Z_{t_{i+1}}^{\mathrm{tgt}}\), which is constructed by re-noising the guided state \(Z_{t_{i+1}}^{\mathrm{edit}}\). However, this guided state would itself depend on \(\hat Z_0^{\mathrm{tgt}}(t_{i+1})\).

\paragraph{One-step delayed approximation.}
RIDGE breaks this cycle by using the most recent available prediction, \(\hat Z_0^{\mathrm{tgt}}(t_i)\), to guide the provisional edited state \(\widetilde Z_{t_{i+1}}^{\mathrm{edit}}\). Since predictions at adjacent timesteps approximate the same underlying clean target state, \(\hat Z_0^{\mathrm{tgt}}(t_i)\) serves as a one-step delayed approximation for the update at \(t_{i+1}\). This design avoids the same-timestep dependency while retaining clean target state guidance during the early high-noise steps.

\section{Full RIDGE Algorithm}
\label{app:full-algorithm}

Algorithm~\ref{alg:ridge_full} gives the complete inference procedure, including the explicit dynamic-mask construction and the optional target-prompt completion stage.

\begin{algorithm}[H]
\caption{Full RIDGE Algorithm}
\label{alg:ridge_full}
\begin{algorithmic}[1]
\Require Clean source state \(Z_0^{\mathrm{src}}\), prompts \(c^{\mathrm{src}}\) and \(c^{\mathrm{tgt}}\), reverse-time schedule \(1=t_0>\cdots>t_N=0\), editing window \((t_{\min},t_{\max}]\), guidance window \([t_{\mathrm{end}},t_{\mathrm{start}}]\), velocity predictor \(v_\theta\).
\Ensure Clean edited state \(Z_0^{\mathrm{edit}}\).

\Statex \textit{// Initialization}
\State Let \(i_{\mathrm{start}}\) and \(i_{\mathrm{end}}\) satisfy \(t_{i_{\mathrm{start}}}=t_{\max}\) and \(t_{i_{\mathrm{end}}}=t_{\min}\)
\State \(Z_{t_{i_{\mathrm{start}}}}^{\mathrm{edit}}\gets Z_0^{\mathrm{src}}\)

\Statex \textit{// Main editing loop}
\For{\(i=i_{\mathrm{start}},\ldots,i_{\mathrm{end}}-1\)}
    \State Sample \(\eta_{t_i}\sim\mathcal{N}(0,I)\)

    \Statex \textit{// Re-noising construction}
    \State \(Z_{t_i}^{\mathrm{src}}\gets (1-t_i)Z_0^{\mathrm{src}}+t_i\eta_{t_i}\)
    \State \(Z_{t_i}^{\mathrm{tgt}}\gets (1-t_i)Z_{t_i}^{\mathrm{edit}}+t_i\eta_{t_i}\)

    \State \(\Delta v_{t_i}\gets
    v_\theta(Z_{t_i}^{\mathrm{tgt}},c^{\mathrm{tgt}},t_i)
    -
    v_\theta(Z_{t_i}^{\mathrm{src}},c^{\mathrm{src}},t_i)\)

    \State \(\widetilde Z_{t_{i+1}}^{\mathrm{edit}}\gets
    Z_{t_i}^{\mathrm{edit}}+(t_{i+1}-t_i)\Delta v_{t_i}\)

    \Statex \textit{// Internal dynamic guidance}
    \If{\(t_{\mathrm{end}}\le t_i\le t_{\mathrm{start}}\)}
        \State \(\hat Z_0^{\mathrm{tgt}}(t_i)\gets
        Z_{t_i}^{\mathrm{tgt}}
        -
        t_i v_\theta(Z_{t_i}^{\mathrm{tgt}},c^{\mathrm{tgt}},t_i)\)

        \State \(D_{t_i}\gets
        \frac{1}{C}\sum_{c=1}^{C}
        \left|
        \hat Z_{0,c}^{\mathrm{tgt}}(t_i)
        -
        Z_{0,c}^{\mathrm{src}}
        \right|\)

        \State \(\phi_{t_i}\gets
        \operatorname{Quantile}_{q}(D_{t_i})\)

        \State \(m_{t_i}\gets
        \sigma\!\left(
        \frac{D_{t_i}-\phi_{t_i}}{\tau}
        \right)\)

        \State \(\gamma_{t_i}\gets
        \lambda_{\mathrm{guide}}/
        (1-\beta_{\mathrm{guide}}t_i)\)

        \State \(Z_{t_{i+1}}^{\mathrm{edit}}\gets
        \widetilde Z_{t_{i+1}}^{\mathrm{edit}}
        +
        \gamma_{t_i}m_{t_i}\odot
        \left(
        \hat Z_0^{\mathrm{tgt}}(t_i)
        -
        \widetilde Z_{t_{i+1}}^{\mathrm{edit}}
        \right)\)
    \Else
        \State \(Z_{t_{i+1}}^{\mathrm{edit}}\gets
        \widetilde Z_{t_{i+1}}^{\mathrm{edit}}\)
    \EndIf
\EndFor

\Statex \textit{// Target-prompt completion}
\If{\(t_{i_{\mathrm{end}}}=0\)}
    \State \Return \(Z_0^{\mathrm{edit}}=Z_{t_{i_{\mathrm{end}}}}^{\mathrm{edit}}\)
\Else
    \State Sample \(\eta_{\mathrm{end}}\sim\mathcal{N}(0,I)\)
    \State \(Y_{t_{i_{\mathrm{end}}}}\gets
    (1-t_{i_{\mathrm{end}}})Z_{t_{i_{\mathrm{end}}}}^{\mathrm{edit}}
    +
    t_{i_{\mathrm{end}}}\eta_{\mathrm{end}}\)

    \For{\(i=i_{\mathrm{end}},\ldots,N-1\)}
        \State \(Y_{t_{i+1}}\gets
        Y_{t_i}
        +
        (t_{i+1}-t_i)
        v_\theta(Y_{t_i},c^{\mathrm{tgt}},t_i)\)
    \EndFor

    \State \Return \(Z_0^{\mathrm{edit}}=Y_{t_N}\)
\EndIf
\end{algorithmic}
\end{algorithm}
\FloatBarrier

\section{Experimental and Evaluation Details}
\label{app:exp-details}

\subsection{Experimental Configuration}

The main experiments use Stable Diffusion 3 Medium (SD3 Medium)~\citep{esser2024scalingrectifiedflowtransformers} as the common backbone. Additional cross-backbone experiments use FLUX.1-dev~\citep{labs2025flux1kontextflowmatching}. Input images are cropped so that both spatial dimensions are divisible by 16 while preserving their native resolution; no fixed-size resizing is applied.

\paragraph{Datasets.}
For PIE-Bench Official~\citep{ju2024pnpinversion} and PIE-Bench++~\citep{huang2024paralleledits}, we evaluate all \textbf{700} edits specified by their official evaluation lists, covering ten edit categories without additional filtering.

\paragraph{Backbones and compared methods.}
Under SD3 Medium, we evaluate RF-Inversion~\citep{rout2024semanticimageinversionediting}, UniEdit~\citep{jiao2026unieditflowunleashinginversionediting}, FlowEdit~\citep{kulikov2025floweditinversionfreetextbasedediting}, FlowAlign~\citep{kim2025flowaligntrajectoryregularizedinversionfreeflowbased}, DRFS~\citep{beaudouin2026deltarectifiedflowsampling}, and RIDGE on both benchmarks. Under FLUX.1-dev, we evaluate the methods for which compatible implementations are available: RF-Inversion, UniEdit, FlowEdit, and RIDGE, using the same evaluation samples.

\paragraph{Compute and software.}
All experiments are run on NVIDIA GeForce RTX 3090 GPUs with 24\,GB of memory. The complete software environment is provided in the accompanying code package.

\subsection{Per-Method Configuration under SD3 Medium}

Tables~\ref{tab:supp_sd3_sampling}--\ref{tab:supp_sd3_schedule} summarize the sampling setup, prompt-conditioning and guidance settings, and method-specific schedules under SD3 Medium. Throughout the experimental configuration, \(t_i\) denotes the rectified-flow timestep, while \(n\) denotes the corresponding discrete reverse-time implementation step.

\begin{table}[H]
\centering
\footnotesize
\caption{Sampling and numerical setup under SD3 Medium. Steps denote the full scheduler trajectory.}
\label{tab:supp_sd3_sampling}
\setlength{\tabcolsep}{3pt}
\begin{tabular}{@{}lcccc@{}}
\toprule
Method & Steps & Scheduler & Seed & Precision \\
\midrule
RF-Inversion & 28 & SD3 FlowMatch & 42 & bf16 \\
UniEdit & 30 & Uni Euler & 8 & fp16 \\
FlowEdit & 50 & SD3 FlowMatch & 42 & fp16 \\
FlowAlign & 33 & FlowAlign Euler & 123 & fp16 \\
DRFS & 50 & SD3 descending & 41 & fp16 \\
RIDGE & 50 & SD3 FlowMatch & 42 & fp16 \\
\bottomrule
\end{tabular}
\end{table}
\FloatBarrier

\begin{table}[H]
\centering
\footnotesize
\caption{Prompt-conditioning and guidance settings under SD3 Medium. The entries follow each method's native guidance formulation and therefore do not always correspond to conventional classifier-free guidance scales.}
\label{tab:supp_sd3_cfg}
\setlength{\tabcolsep}{2pt}
\begin{tabular}{@{}l>{\raggedright\arraybackslash}p{0.34\linewidth}>{\raggedright\arraybackslash}p{0.34\linewidth}@{}}
\toprule
Method & Source-side guidance & Target/edit guidance \\
\midrule
RF-Inversion & Null-conditioned inversion & Target CFG \(3.5\) \\
UniEdit & Source scale \(1.0\) & Target scale \(1.0\) \\
FlowEdit & Source CFG \(3.5\) & Target CFG \(13.5\) \\
FlowAlign & Direct source conditioning & Source-to-target scale \(13.5\) \\
DRFS & Source CFG \(6.0\) & Target CFG \(16.5\) \\
RIDGE & Source CFG \(3.5\) & Target CFG \(13.5\) \\
\bottomrule
\end{tabular}
\end{table}
\FloatBarrier

\noindent
For FlowAlign, direct source conditioning means that the source prompt is used to predict the conditional velocity without a separate conventional source CFG scale. Its target-side scale extrapolates from the source-conditioned velocity toward the target-conditioned velocity.

\begin{table}[H]
\centering
\small
\caption{Method-specific update schedules and additional settings under SD3 Medium.}
\label{tab:supp_sd3_schedule}
\setlength{\tabcolsep}{4pt}
\begin{tabular}{@{}l>{\raggedright\arraybackslash}p{\dimexpr\linewidth-2.2cm\relax}@{}}
\toprule
Method & Setting \\
\midrule
RF-Inversion &
Uses 28 inversion and denoising steps. Edit interpolation is applied at steps \(0,\ldots,8\), with \(\gamma=0.5\), \(\eta_{\mathrm{base}}=0.95\), and a constant \(\eta\) schedule. \\
UniEdit &
Uses 15 inversion steps followed by 15 editing steps, with \(\alpha=0.6\) and \(\omega=5.0\). \\
FlowEdit &
Uses \(T_{\mathrm{steps}}=50\), \(n_{\max}=36\), and \(n_{\min}=5\). Velocity-difference editing is applied for \(n=36,\ldots,6\), followed by tail denoising for \(n=5,\ldots,1\). One fresh Gaussian noise sample is drawn at every editing step, corresponding to \(n_{\mathrm{avg}}=1\). \\
FlowAlign &
Uses \(N_{\mathrm{FE}}=33\); all 33 reverse-time steps participate in editing. \\
DRFS &
Uses \(T_{\mathrm{steps}}=50\) descending latent-optimization steps, \(\eta=1.0\), SGD with batch size \(B=1\), and the released \texttt{lr="custom"} schedule without modification. \\
RIDGE &
Uses the same editable interval, tail-denoising schedule, and per-step fresh-noise sampling as FlowEdit. RIDGE-specific guidance settings are reported in Table~\ref{tab:supp_ridge_guidance}. \\
\bottomrule
\end{tabular}
\end{table}
\FloatBarrier

\subsection{Per-Method Configuration under FLUX.1-dev}

Tables~\ref{tab:supp_flux_sampling}--\ref{tab:supp_flux_schedule} summarize the sampling setup, prompt-guidance settings, and method-specific schedules under FLUX.1-dev. FlowEdit and RIDGE use sequential CPU offload.

\begin{table}[H]
\centering
\footnotesize
\caption{Sampling and numerical setup under FLUX.1-dev. Steps denote the full scheduler trajectory.}
\label{tab:supp_flux_sampling}
\setlength{\tabcolsep}{3pt}
\begin{tabular}{@{}lcccc@{}}
\toprule
Method & Steps & Scheduler & Seed & Precision \\
\midrule
RF-Inversion & 28 & FLUX FlowMatch Euler & 42 & bf16 \\
UniEdit & 30 & Uni Euler & 8 & bf16 \\
FlowEdit & 28 & FLUX FlowMatch Euler & 42 & fp16 \\
RIDGE & 28 & FLUX FlowMatch Euler & 42 & fp16 \\
\bottomrule
\end{tabular}
\end{table}
\FloatBarrier

\begin{table}[H]
\centering
\small
\caption{Prompt-guidance settings under FLUX.1-dev. A dash indicates that no separate source-side guidance scale is defined by the released implementation; it does not imply that the source prompt is unused.}
\label{tab:supp_flux_cfg}
\setlength{\tabcolsep}{4pt}
\begin{tabular}{@{}lcc@{}}
\toprule
Method & Source-side scale & Target/edit scale \\
\midrule
RF-Inversion & --- & 1.0 \\
UniEdit & 1.0 & 1.0 \\
FlowEdit & 1.5 & 5.5 \\
RIDGE & 1.5 & 5.5 \\
\bottomrule
\end{tabular}
\end{table}
\FloatBarrier

\begin{table}[H]
\centering
\small
\caption{Method-specific update schedules and additional settings under FLUX.1-dev.}
\label{tab:supp_flux_schedule}
\setlength{\tabcolsep}{4pt}
\begin{tabular}{@{}l>{\raggedright\arraybackslash}p{\dimexpr\linewidth-2.2cm\relax}@{}}
\toprule
Method & Setting \\
\midrule
RF-Inversion &
Uses 28 inversion and denoising steps, with \(\eta=0.9\) and \(\gamma=0.5\). Edit interpolation is applied over the normalized interval \([0.0,0.25]\). \\
UniEdit &
Uses 15 inversion steps followed by 15 editing steps, with \(\alpha=0.6\), \(\omega=5.0\), and inversion and editing guidance scales of \(1.0\). \\
FlowEdit &
Uses \(T_{\mathrm{steps}}=28\), \(n_{\max}=24\), \(n_{\min}=0\), and \(n_{\mathrm{avg}}=1\). Editing is applied for \(n=24,\ldots,1\). \\
RIDGE &
Uses the same editable interval and per-step fresh-noise sampling as FlowEdit. RIDGE-specific guidance settings are reported in Table~\ref{tab:supp_ridge_guidance}. \\
\bottomrule
\end{tabular}
\end{table}
\FloatBarrier

\subsection{RIDGE-Specific Guidance Configuration}

For controlled comparisons, RIDGE matches FlowEdit on the scheduler, editable interval, prompt-guidance settings, random seed, numerical precision, and per-step fresh-noise sampling under each backbone. Table~\ref{tab:supp_ridge_guidance} reports only the RIDGE-specific reverse-time guidance window and \(\lambda_{\mathrm{guide}}\). The listed window specifies the steps at which both the soft dynamic mask and the guidance update are applied.

\begin{table}[H]
\centering
\small
\caption{RIDGE-specific guidance settings by backbone.
In both cases, \(\beta_{\mathrm{guide}}=0.02\), mask quantile \(q=0.7\), and temperature \(\tau=0.2\).}
\label{tab:supp_ridge_guidance}
\setlength{\tabcolsep}{6pt}
\begin{tabular}{@{}lcc@{}}
\toprule
Backbone & Guidance window & \(\lambda_{\mathrm{guide}}\) \\
\midrule
SD3 Medium & \(n=36,\ldots,30\) & 0.012 \\
FLUX.1-dev & \(n=24,\ldots,18\) & 0.024 \\
\bottomrule
\end{tabular}
\end{table}
\FloatBarrier

\subsection{Baseline Implementations}

We use publicly released implementations for all baselines and retain their method-specific configurations without additional tuning. The repositories and revisions used in our experiments are listed below:
\begin{itemize}
    \item FlowEdit~\citep{kulikov2025floweditinversionfreetextbasedediting}: \url{https://github.com/fallenshock/FlowEdit}, commit \texttt{9a2e05e}.
    \item FlowAlign~\citep{kim2025flowaligntrajectoryregularizedinversionfreeflowbased}: \url{https://github.com/FlowAlign/FlowAlign}, commit \texttt{29c6fe1}.
    \item DRFS~\citep{beaudouin2026deltarectifiedflowsampling}: \url{https://github.com/Harvard-AI-and-Robotics-Lab/DeltaVelocityRectifiedFlow}, commit \texttt{c59ce9d}.
    \item UniEdit~\citep{jiao2026unieditflowunleashinginversionediting}: \url{https://github.com/DSL-Lab/UniEdit-Flow}, commit \texttt{2a5385e}.
    \item RF-Inversion under SD3 Medium~\citep{rout2024semanticimageinversionediting}: the public SD3 port at \url{https://github.com/pureexe/rf-inversion-sd3}, commit \texttt{0d303d3}.
    \item RF-Inversion under FLUX.1-dev~\citep{rout2024semanticimageinversionediting}: the public Diffusers community pipeline at \url{https://github.com/huggingface/diffusers/blob/c9e4fab/examples/community/pipeline_flux_rf_inversion.py}, commit \texttt{c9e4fab}.
    \item PIE-Bench evaluation code~\citep{ju2024pnpinversion}: \url{https://github.com/cure-lab/PnPInversion}, commit \texttt{07f97f4}.
\end{itemize}

Baseline algorithms and method-specific hyperparameters are unchanged. All methods use the same benchmark preprocessing protocol, which preserves the native image resolution while cropping both spatial dimensions to multiples of 16. For RF-Inversion under FLUX.1-dev, we make only tensor-reshaping compatibility adjustments required by our Diffusers environment; these adjustments do not change the algorithmic operations or hyperparameters. The exact experimental configurations used for all methods are reported in the tables above.

\subsection{Metric Implementation Details}
\label{sec:metric_impl}

\paragraph{Structural preservation.}
We compute StructDist following the official PIE-Bench evaluation protocol~\citep{ju2024pnpinversion}. DINO ViT-B/8~\citep{caron2021emergingpropertiesselfsupervisedvision} extracts key embeddings from the final self-attention block. After concatenating the attention heads for each token, including the CLS token, we construct a token-wise self-similarity matrix using pairwise cosine similarities. StructDist is the mean-squared discrepancy between the self-similarity matrices of the source and edited images~\citep{tumanyan2022splicing}. Lower values indicate better structural preservation.

\paragraph{Unedited-region fidelity.}
PSNR$_{\rm u}$ and LPIPS$_{\rm u}$ are computed between the source and edited images over the complement of the annotated editing region. Following the official PIE-Bench implementation, PSNR uses a data range of \(1\), and LPIPS uses the SqueezeNet backbone~\citep{zhang2018unreasonableeffectivenessdeepfeatures}. Higher PSNR$_{\rm u}$ and lower LPIPS$_{\rm u}$ indicate better preservation.

\paragraph{Target alignment.}
CLIP$_{\rm tgt}$ measures alignment between the edited image and its target prompt. Across all benchmark--backbone combinations, we use the official PIE-Bench evaluator~\citep{ju2024pnpinversion}, implemented with TorchMetrics \texttt{CLIPScore} and the \texttt{openai/\allowbreak clip-vit-\allowbreak large-patch14} checkpoint~\citep{radford2021learningtransferablevisualmodels}. The metric is applied to the full edited image and target prompt using the evaluator's default preprocessing. Higher values indicate stronger target-prompt alignment.

\paragraph{Global consistency.}
Let \(f_{\mathrm{cls}}\) denote the final CLS embedding produced by DINOv2-Base~\citep{oquab2024dinov2learningrobustvisual}. We use the \texttt{facebook/dinov2-base} checkpoint and define the distance between a source image \(I_{\mathrm{src}}\) and edited image \(I_{\mathrm{edit}}\) as
\begin{equation}
D_{\mathrm{DINO}}
=
1-
\frac{
\left\langle
f_{\mathrm{cls}}(I_{\mathrm{src}}),
f_{\mathrm{cls}}(I_{\mathrm{edit}})
\right\rangle
}{
\left\|
f_{\mathrm{cls}}(I_{\mathrm{src}})
\right\|_2
\left\|
f_{\mathrm{cls}}(I_{\mathrm{edit}})
\right\|_2
}.
\label{eq:supp_dino_distance}
\end{equation}
Lower values indicate greater global visual consistency. Unlike StructDist, which compares token-level DINO self-similarity matrices to assess structural consistency, this metric compares full-image CLS embeddings. Prior editing evaluation has similarly used full-image DINO features to compare input and edited images~\citep{sheynin2024emuedit}; we use DINOv2 CLS features and report cosine distance rather than similarity. Images are processed independently using the checkpoint's Hugging Face image processor.

\paragraph{Aesthetic quality.}
Aes. is computed by applying the improved LAION aesthetic predictor MLP~\citep{schuhmann2022improvedaesthetic} to \(\ell_2\)-normalized OpenAI CLIP ViT-L/14 image features~\citep{radford2021learningtransferablevisualmodels}. We use the checkpoint \texttt{sac+logos+ava1-\allowbreak l14-\allowbreak linearMSE.pth}. Higher values indicate higher predicted aesthetic quality.

\paragraph{Dataset-level aggregation.}
Each dataset-level metric is the arithmetic mean over samples with finite values. Non-finite sample values are excluded for the corresponding metric, and the aggregate is set to NaN if no finite value remains.

\paragraph{Average score.}
Avg Score provides an additional summary of the overall performance balance within each table. Let \(x_{j,m}\) denote the dataset-level result of method \(j\) on metric \(m\), and let \(x_m^{\min}\) and \(x_m^{\max}\) denote the minimum and maximum values across the methods compared in that table. The normalized score is
\begin{equation}
s_{j,m}
=
\begin{cases}
\dfrac{x_{j,m}-x_m^{\min}}
{x_m^{\max}-x_m^{\min}},
& \text{if higher is better}, \\[8pt]
\dfrac{x_m^{\max}-x_{j,m}}
{x_m^{\max}-x_m^{\min}},
& \text{if lower is better}.
\end{cases}
\label{eq:supp_metric_normalization}
\end{equation}
If \(x_m^{\max}=x_m^{\min}\), all methods receive \(s_{j,m}=1\) for that metric. Avg Score is then computed as
\begin{equation}
\operatorname{AvgScore}_j
=
\frac{1}{|\mathcal{M}|}
\sum_{m\in\mathcal{M}} s_{j,m},
\label{eq:supp_average_score}
\end{equation}
where \(\mathcal{M}\) contains StructDist, PSNR$_{\rm u}$, LPIPS$_{\rm u}$, CLIP$_{\rm tgt}$, DINO distance, and Aes. Normalization is performed separately within each dataset and comparison table. All dataset-level values used in the reported tables are finite.

\section{Additional Quantitative Results}
\label{app:add-quant}
\label{sec:flux_results}

Table~\ref{tab:flux_selected_metrics} reports quantitative results under FLUX.1-dev on PIE-Bench Official and PIE-Bench++. RIDGE consistently achieves the lowest StructDist, LPIPS, and DINO distance on both benchmarks, demonstrating strong source preservation and global visual consistency under a different backbone. These results further support the cross-backbone robustness of RIDGE.

\begin{table}[H]
\centering
\caption{Quantitative comparison under FLUX.1-dev on PIE-Bench Official and PIE-Bench++.
RF-Inv. denotes RF-Inversion. PSNR and LPIPS are computed in unedited regions.}
\label{tab:flux_selected_metrics}
{\small
\setlength{\tabcolsep}{1.5pt}
\begin{tabular}{@{}cllccccccc@{}}
\toprule
\multirow{2}{*}{Dataset} & \multirow{2}{*}{Type} & \multirow{2}{*}{Method}
& \multicolumn{4}{c}{Source Preservation}
& \multicolumn{1}{c}{Target Alignment}
& \multicolumn{1}{c}{Aesthetic}
& \multirow{2}{*}{Avg Sc.$\uparrow$} \\
\cmidrule(lr){4-7}\cmidrule(lr){8-8}\cmidrule(lr){9-9}
& & & Struct.$\downarrow$ & PSNR$\uparrow$ & LPIPS$\downarrow$ & DINO$\downarrow$ & CLIP$_{\rm tgt}\uparrow$ & Aes.$\uparrow$ & \\
\midrule
\multirow{4}{*}{PIE-Bench} & Inv-B & RF-Inv.
& 0.037 & 21.025 & 0.167 & 0.372 & 23.878 & 5.538 & 0.000 \\
& Inv-B & UniEdit
& 0.011 & 29.012 & 0.063 & 0.224 & 25.837 & 5.682 & 0.773 \\
& Inv-F & FlowEdit
& 0.028 & 21.787 & 0.112 & 0.326 & 26.312 & 5.871 & 0.511 \\
\oursrow
& Inv-F & \textbf{RIDGE}
& 0.010 & 26.145 & 0.047 & 0.113 & 24.644 & 5.801 & 0.791 \\
\midrule
\multirow{4}{*}{PIE-Bench++} & Inv-B & RF-Inv.
& 0.037 & 21.624 & 0.154 & 0.408 & 23.193 & 5.521 & 0.000 \\
& Inv-B & UniEdit
& 0.014 & 29.083 & 0.064 & 0.301 & 25.565 & 5.649 & 0.701 \\
& Inv-F & FlowEdit
& 0.031 & 22.288 & 0.103 & 0.388 & 26.181 & 5.889 & 0.475 \\
\oursrow
& Inv-F & \textbf{RIDGE}
& 0.010 & 26.796 & 0.044 & 0.133 & 23.384 & 5.804 & 0.754 \\
\bottomrule
\end{tabular}
}
\end{table}
\FloatBarrier

\section{Human Preference Study}
\label{app:user-study}

\subsection{Setting}

\begin{figure}[H]
\centering
\includegraphics[width=0.62\linewidth]{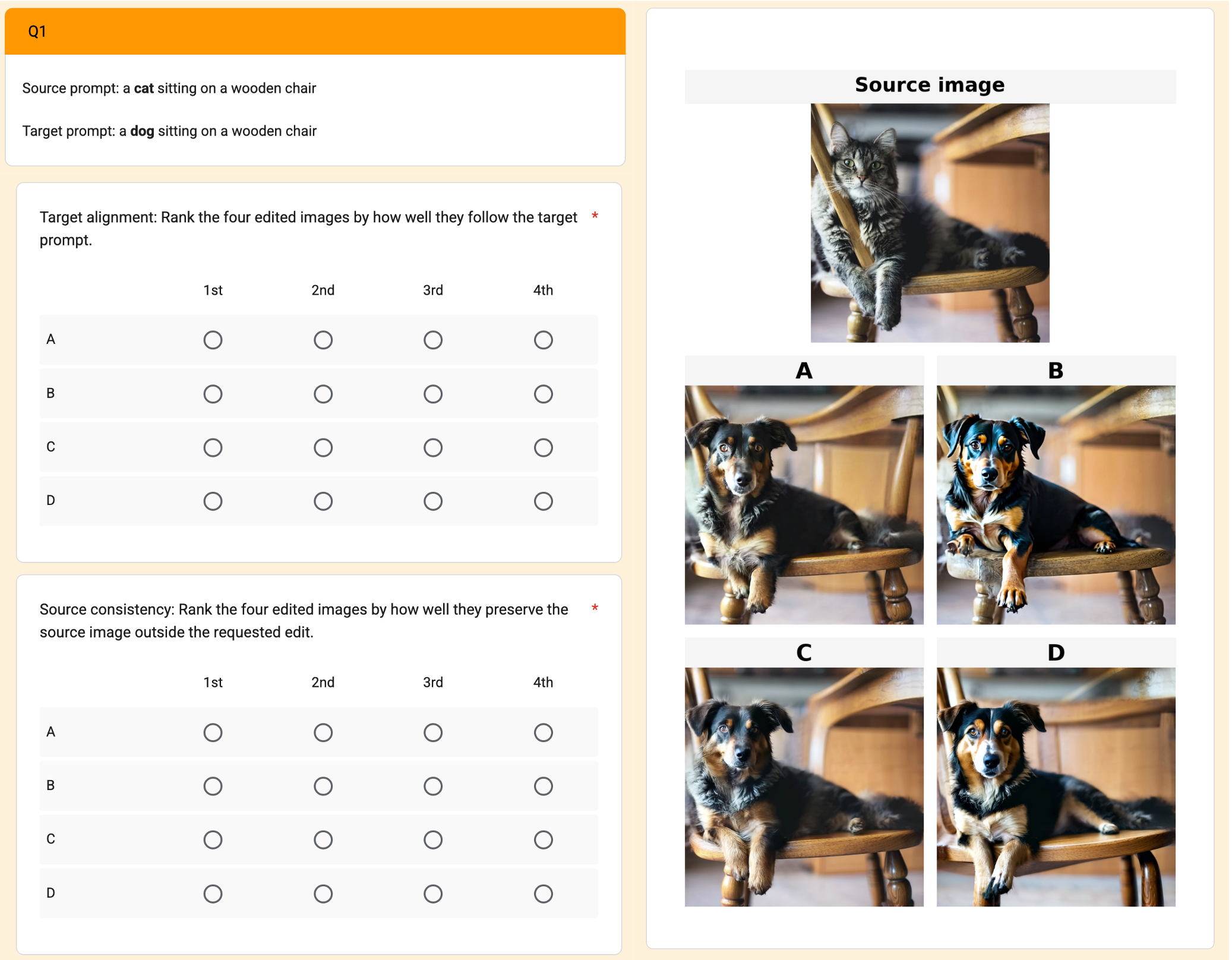}
\caption{Example question from the human preference study interface. Participants rank anonymized edits A--D under target alignment and source consistency.}
\label{fig:user_study_demo}
\end{figure}

We conduct a human preference study comparing FlowEdit~\citep{kulikov2025floweditinversionfreetextbasedediting}, FlowAlign~\citep{kim2025flowaligntrajectoryregularizedinversionfreeflowbased}, DRFS~\citep{beaudouin2026deltarectifiedflowsampling}, and RIDGE on 20 editing cases sampled from PIE-Bench Official and PIE-Bench++. For each case, participants are shown the source image, source and target prompts, and four anonymized outputs labeled A--D, with label assignments randomized across cases. They rank the outputs from first to fourth for \emph{target alignment} and \emph{source consistency}, which respectively assess fulfillment of the requested edit and preservation of unrelated source content. Fig.~\ref{fig:user_study_demo} shows an example question.

\FloatBarrier

We collected 15 complete responses, resulting in \(15\times20=300\) rank observations for each method under each criterion. We report the average rank, where lower is better, and the top-1 rate, which measures how often a method is ranked first.

\subsection{Results}

As shown in Table~\ref{tab:user_study_ranking}, RIDGE achieves the best average rank and the highest top-1 rate under both criteria. It is ranked first in \(58.0\%\) of the target-alignment judgments and \(68.7\%\) of the source-consistency judgments. When aggregating both criteria, RIDGE is ranked first in \(380\) out of \(600\) method-level rankings, corresponding to a top-1 rate of \(63.3\%\).

\begin{table}[H]
\centering
\small
\setlength{\tabcolsep}{5pt}
\caption{Human preference study results over 15 complete responses and 20 editing trials. Participants rank all four methods under target alignment and source consistency.}
\label{tab:user_study_ranking}
\begin{tabular}{@{}lcccccc@{}}
\toprule
\multirow{2}{*}{Method}
& \multicolumn{2}{c}{Target Alignment}
& \multicolumn{2}{c}{Source Consistency}
& \multicolumn{2}{c}{Combined} \\
\cmidrule(lr){2-3}\cmidrule(lr){4-5}\cmidrule(lr){6-7}
& Avg. rank \(\downarrow\) & Top-1 \(\uparrow\)
& Avg. rank \(\downarrow\) & Top-1 \(\uparrow\)
& Avg. rank \(\downarrow\) & Top-1 \(\uparrow\) \\
\midrule
FlowEdit  & 2.757 & 13.7\% & 3.047 & 7.0\%  & 2.902 & 10.3\% \\
FlowAlign & 3.187 & 11.7\% & 3.040 & 14.7\% & 3.113 & 13.2\% \\
DRFS      & 2.347 & 16.7\% & 2.443 & 9.7\%  & 2.395 & 13.2\% \\
\textbf{RIDGE}
          & \best{1.710} & \best{58.0\%}
          & \best{1.470} & \best{68.7\%}
          & \best{1.590} & \best{63.3\%} \\
\bottomrule
\end{tabular}
\end{table}
\FloatBarrier

The method-to-label assignments are randomized across cases but are not perfectly counterbalanced, so small residual position effects may remain. We therefore interpret the study as descriptive supporting evidence rather than as a population-level estimate.

\section{Clean Target State Prediction Analysis}
\label{app:pred-analysis}

RIDGE uses a clean target state prediction to guide the edited state during the early high-noise steps. Its one-step delayed guidance relies on the most recent available prediction to guide the following update. We examine how these predictions evolve using two complementary latent-space diagnostics over all 700 PIE-Bench Official samples. At each reverse-time step \(n\in\{36,\ldots,30\}\), we record the clean target state prediction \(\hat Z_0^{\mathrm{tgt}}(t_i)\), where \(n\) corresponds to timestep \(t_i\) and the following reverse-time step \(n-1\) corresponds to \(t_{i+1}\). All distances are measured using per-element latent RMSE, where \(D\) denotes the number of latent elements.

\paragraph{Adjacent-prediction distance.}
To measure consistency between predictions at consecutive guidance steps, we define
\begin{equation}
d_{\mathrm{adj}}(n)
=
\frac{1}{\sqrt{D}}
\left\|
\hat Z_0^{\mathrm{tgt}}(t_i)
-
\hat Z_0^{\mathrm{tgt}}(t_{i+1})
\right\|_2.
\label{eq:supp_adjacent_prediction_distance}
\end{equation}
This quantity is reported for \(n=36,\ldots,31\), since \(n=30\) is the final step in the guidance window and has no subsequent prediction. Lower values indicate greater consistency between predictions at adjacent steps.

\paragraph{Distance to the final edited state.}
We additionally measure the distance from each prediction to the final edited state produced by the same RIDGE trajectory:
\begin{equation}
d_{\mathrm{final}}(n)
=
\frac{1}{\sqrt{D}}
\left\|
\hat Z_0^{\mathrm{tgt}}(t_i)
-
Z_0^{\mathrm{edit}}
\right\|_2.
\label{eq:supp_final_prediction_distance}
\end{equation}
Because \(Z_0^{\mathrm{edit}}\) is the endpoint produced by RIDGE rather than a ground-truth clean target state, \(d_{\mathrm{final}}\) measures proximity to the trajectory endpoint and should not be interpreted as prediction accuracy.

\begin{figure}[H]
\centering
\begin{minipage}[t]{0.49\textwidth}
\centering
\includegraphics[width=\linewidth]{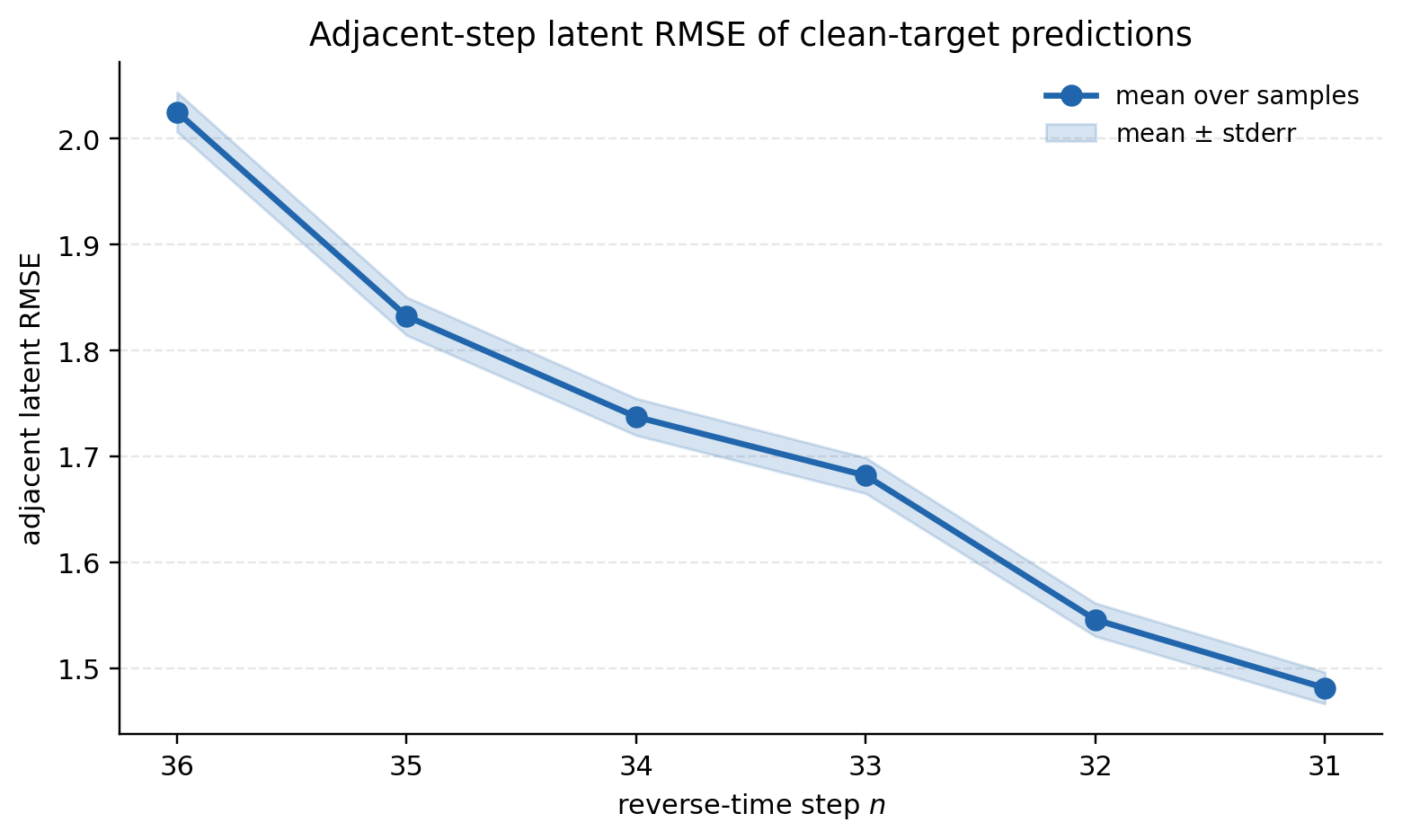}
\end{minipage}
\hfill
\begin{minipage}[t]{0.49\textwidth}
\centering
\includegraphics[width=\linewidth]{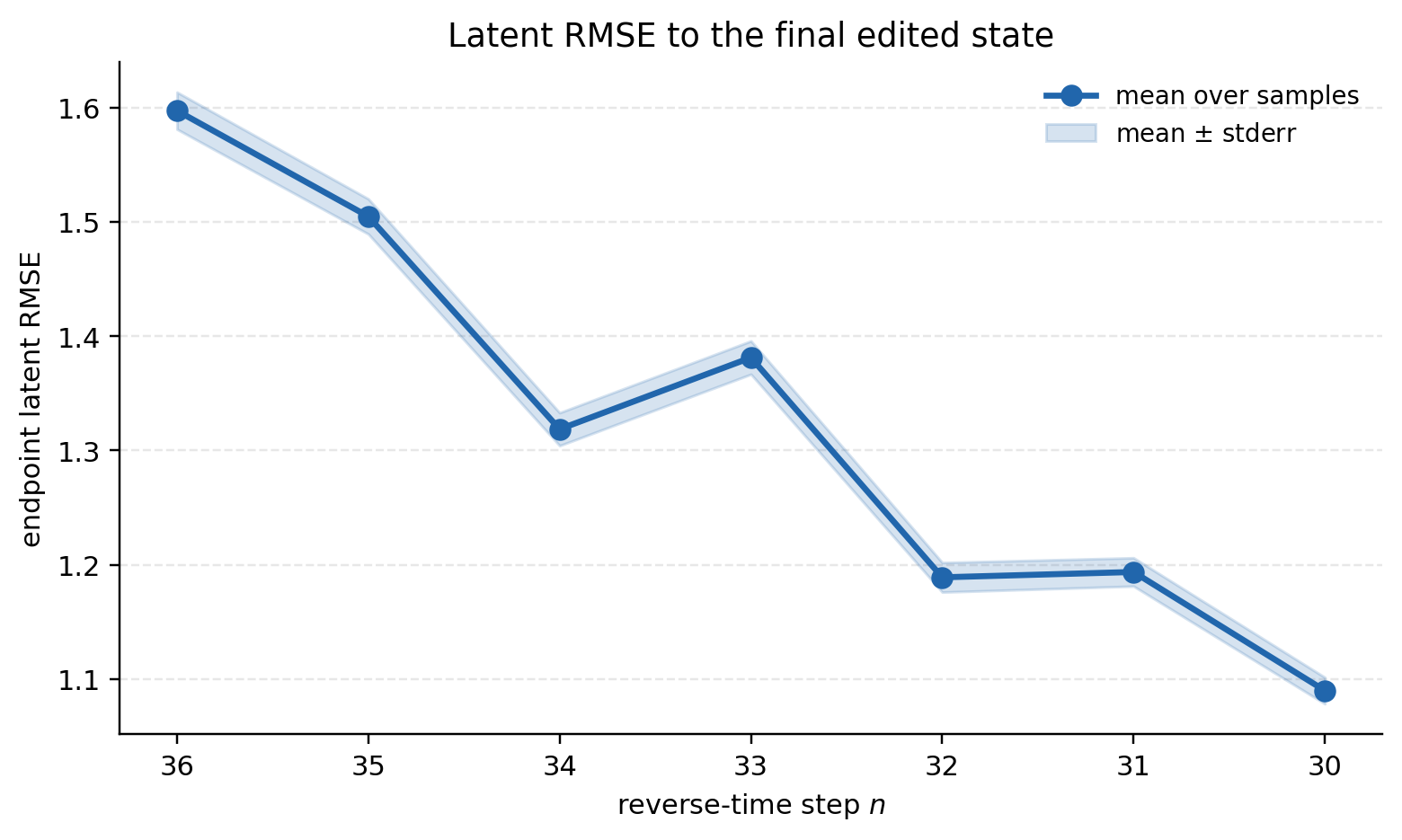}
\end{minipage}
\caption{Clean target state prediction diagnostics over all 700 PIE-Bench Official samples. Left: per-element latent RMSE between predictions at consecutive guidance steps. Right: per-element latent RMSE between each prediction and the final edited state from the same RIDGE trajectory. Curves show the sample means, and shaded regions show the standard errors of the means.}
\label{fig:prediction_diagnostics}
\end{figure}

Figure~\ref{fig:prediction_diagnostics} reports the mean distances over all 700 samples. The shaded regions show the standard errors of the sample means, computed as the sample standard deviation divided by \(\sqrt{700}\). The adjacent-prediction distance decreases consistently toward lower noise levels, indicating that consecutive clean target state predictions become more temporally consistent as editing proceeds. This observation supports using the most recent available prediction as a one-step delayed approximation for the following update.

The distance to the final edited state also exhibits an overall decreasing trend, although it is not strictly monotonic at every step. Later predictions therefore tend to lie closer to the final trajectory endpoint. This diagnostic characterizes how the predictions evolve along the RIDGE trajectory, rather than their accuracy with respect to an unavailable ground-truth target state.

\begin{figure}[H]
\centering
\includegraphics[width=0.62\linewidth]{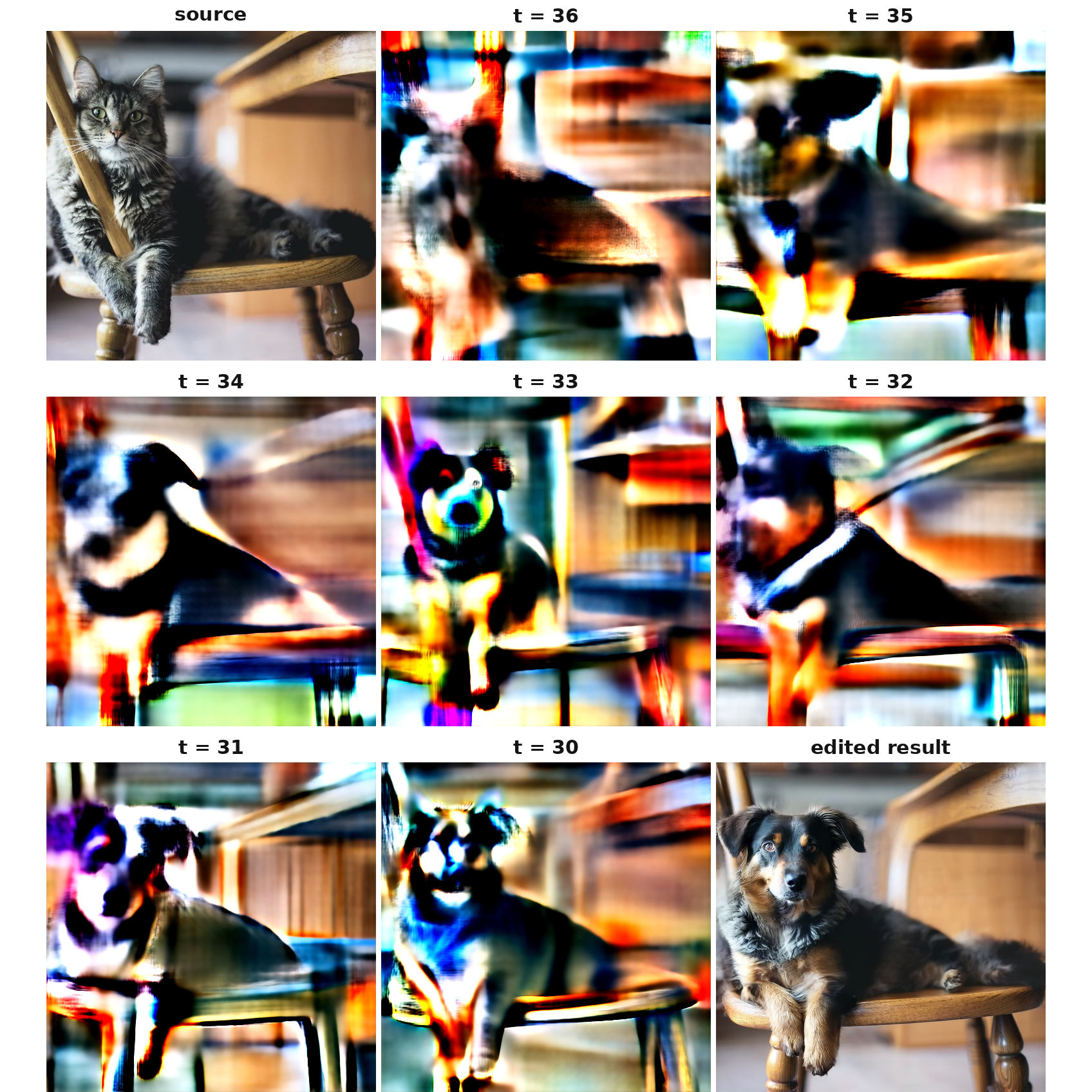}
\caption{Clean target state predictions across guidance steps. The source image and final edited result are shown as references. Example edit: ``a cat sitting on a wooden chair'' \(\rightarrow\) ``a dog sitting on a wooden chair''.}
\label{fig:ablation_pred_vis}
\end{figure}

Figure~\ref{fig:ablation_pred_vis} further visualizes the clean target state predictions for one editing example. In this example, earlier predictions exhibit broader and less coherent spatial changes, whereas later predictions become more consistent with the final edited result. Together with the quantitative trends above, this observation is consistent with the motivation for the soft dynamic mask, which restricts the spatial influence of guidance when early high-noise predictions may contain changes outside the intended edit region.

\FloatBarrier

\FloatBarrier

\section{Additional Ablation Results}
\label{app:add-ablation}

The default SD3 Medium configuration was fixed before the full-benchmark evaluation. The following sweeps are reported as post-hoc sensitivity analyses.

\subsection{Guidance-Window Sweep}

\begin{figure}[H]
\centering
\includegraphics[width=\textwidth]{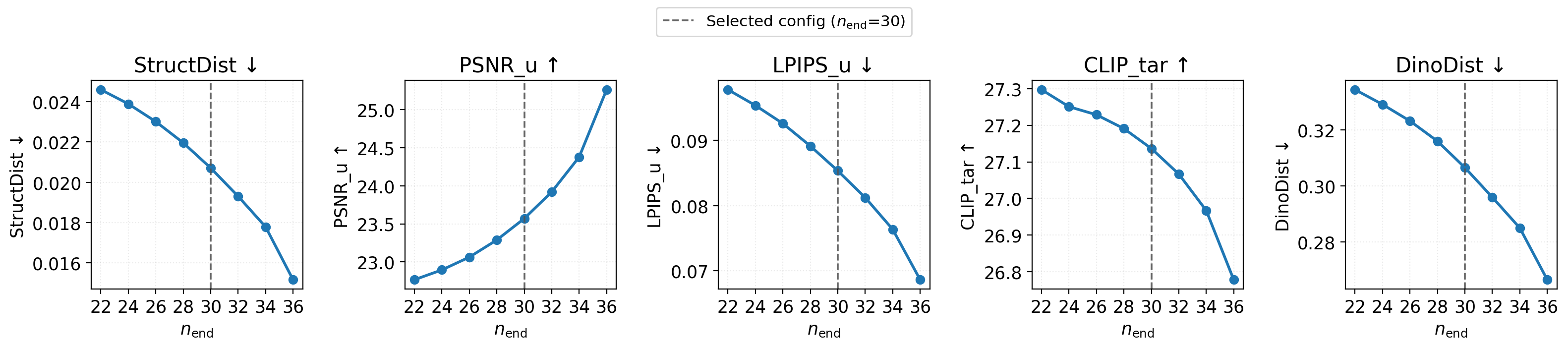}
\caption{Full guidance-window sweep on PIE-Bench Official under SD3 Medium. We fix \(n_{\mathrm{start}}=36\) and vary \(n_{\mathrm{end}}\). The dashed line marks the default setting (\(n_{\mathrm{end}}=30\)).}
\label{fig:ablation_guidance_window_full}
\end{figure}

Fig.~\ref{fig:ablation_guidance_window_full} reports the same guidance-window sweep as in the main text, with additional metrics beyond StructDist and CLIP$_{\rm tgt}$. The same trade-off holds: longer guidance improves target alignment but weakens source preservation and related source-similarity metrics. The default setting \(n_{\mathrm{end}}=30\) lies near the observed trade-off.

\FloatBarrier

\subsection{Guidance-Strength Sweep}

\begin{figure}[H]
\centering
\includegraphics[width=\textwidth]{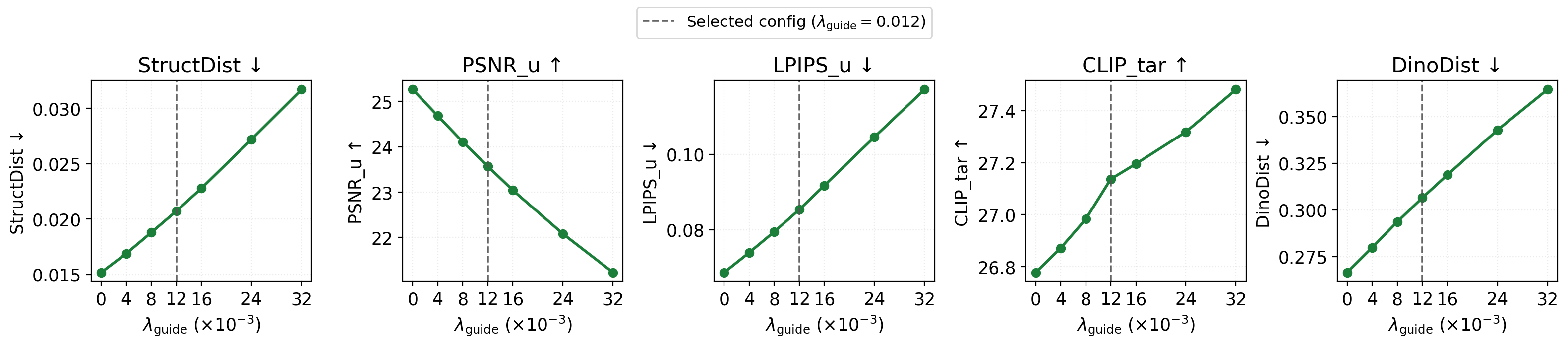}
\caption{Full guidance-strength sweep on PIE-Bench Official under SD3 Medium. We vary \(\lambda_{\mathrm{guide}}\) while keeping the guidance window fixed. The dashed line marks the default setting (\(\lambda_{\mathrm{guide}}=0.012\)).}
\label{fig:ablation_lambda_full}
\end{figure}

Fig.~\ref{fig:ablation_lambda_full} reports the same \(\lambda_{\mathrm{guide}}\) sweep as in the main text, with additional metrics beyond StructDist and CLIP$_{\rm tgt}$. Larger \(\lambda_{\mathrm{guide}}\) improves target alignment but weakens source preservation and related source-similarity metrics. The default setting \(\lambda_{\mathrm{guide}}=0.012\) lies near the observed trade-off.

\FloatBarrier

\end{document}